\documentclass{article}
\PassOptionsToPackage{numbers, sort&compress}{natbib}
\usepackage[preprint]{neurips_2020}

\usepackage[utf8]{inputenc} 
\usepackage[T1]{fontenc}    
\usepackage{hyperref}       
\usepackage{url}            
\usepackage{booktabs}       
\usepackage{amsfonts}       
\usepackage{nicefrac}       
\usepackage{microtype}      

\usepackage{graphicx}
\usepackage{subfigure}
\usepackage{wrapfig}
\usepackage{amsmath}
\usepackage{amssymb}
\usepackage{amsthm}

\usepackage{hyperref}

\newcommand{\indistness}{Affinity}
\newcommand{\cifar}[0]{CIFAR-10}
\newcommand{\imagenet}[0]{ImageNet}
\newcommand{\mnist}[0]{MNIST}
\newcommand{\diversity}{Diversity}

\newcommand{\aug}[1]{{\small{\texttt{#1}}}}
\newtheorem{definition}{Definition}

\title{\indistness{} and \diversity{}: Quantifying Mechanisms of Data Augmentation}
\author{
  Raphael Gontijo-Lopes\thanks{Equal contribution} \\
  Google Brain\\ Mountain View, CA 94043
  \And
  Sylvia J.~Smullin$^*$ \\
  Google \\
  Mountain View, CA 94043 \\
  \AND
  Ekin D.~Cubuk \\
  Google Brain\\ Mountain View, CA 94043\\
  \texttt{cubuk@google.com} \\
  \And
  Ethan Dyer \\
  Google \\ Mountain View, CA 94043\\
  \texttt{edyer@google.com} \\
}

\begin{document}
\maketitle

\begin{abstract}
Though data augmentation has become a standard component of deep neural network training, the underlying mechanism behind the effectiveness of these techniques remains poorly understood. In practice, augmentation policies are often chosen using heuristics of either distribution shift or augmentation diversity. Inspired by these, we seek to quantify how data augmentation improves model generalization. To this end, we introduce interpretable and easy-to-compute measures: \indistness{} and \diversity{}. We find that augmentation performance is predicted not by either of these alone but by jointly optimizing the two. 
\end{abstract}
\begin{figure}[ht]
\vskip 0.1in 
\begin{center}
\centerline{%
    \hfill%
    \begin{subfigure}[\indistness{} vs \diversity{}]{%
        \includegraphics[width=0.41\columnwidth]{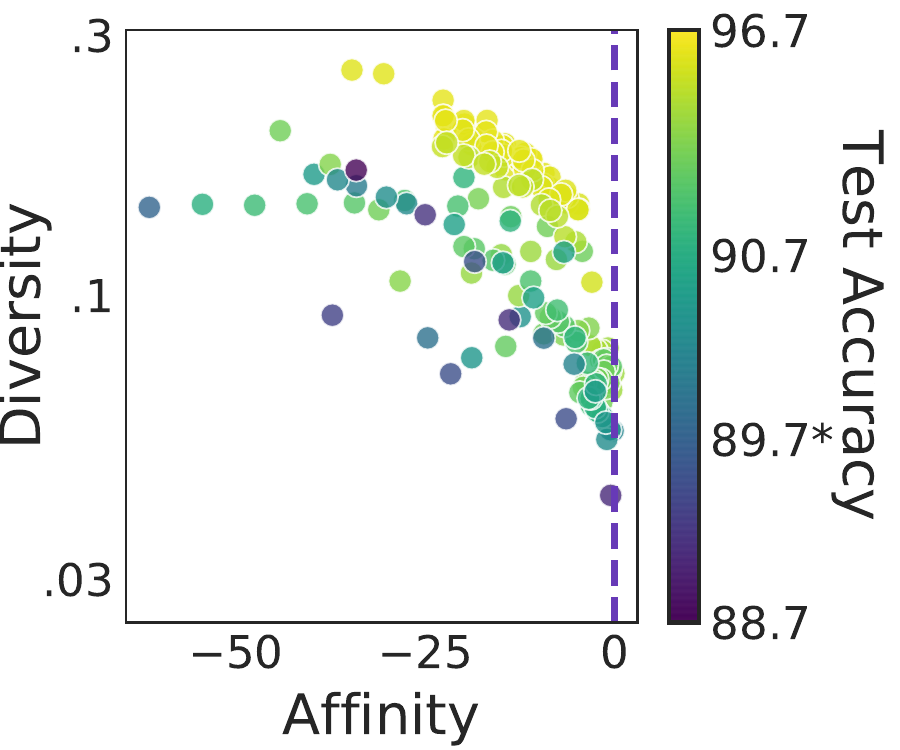}%
    }\end{subfigure}%
    \hfill%
    \begin{subfigure}[Model's View of Data]{%
        \includegraphics[width=0.41\columnwidth]{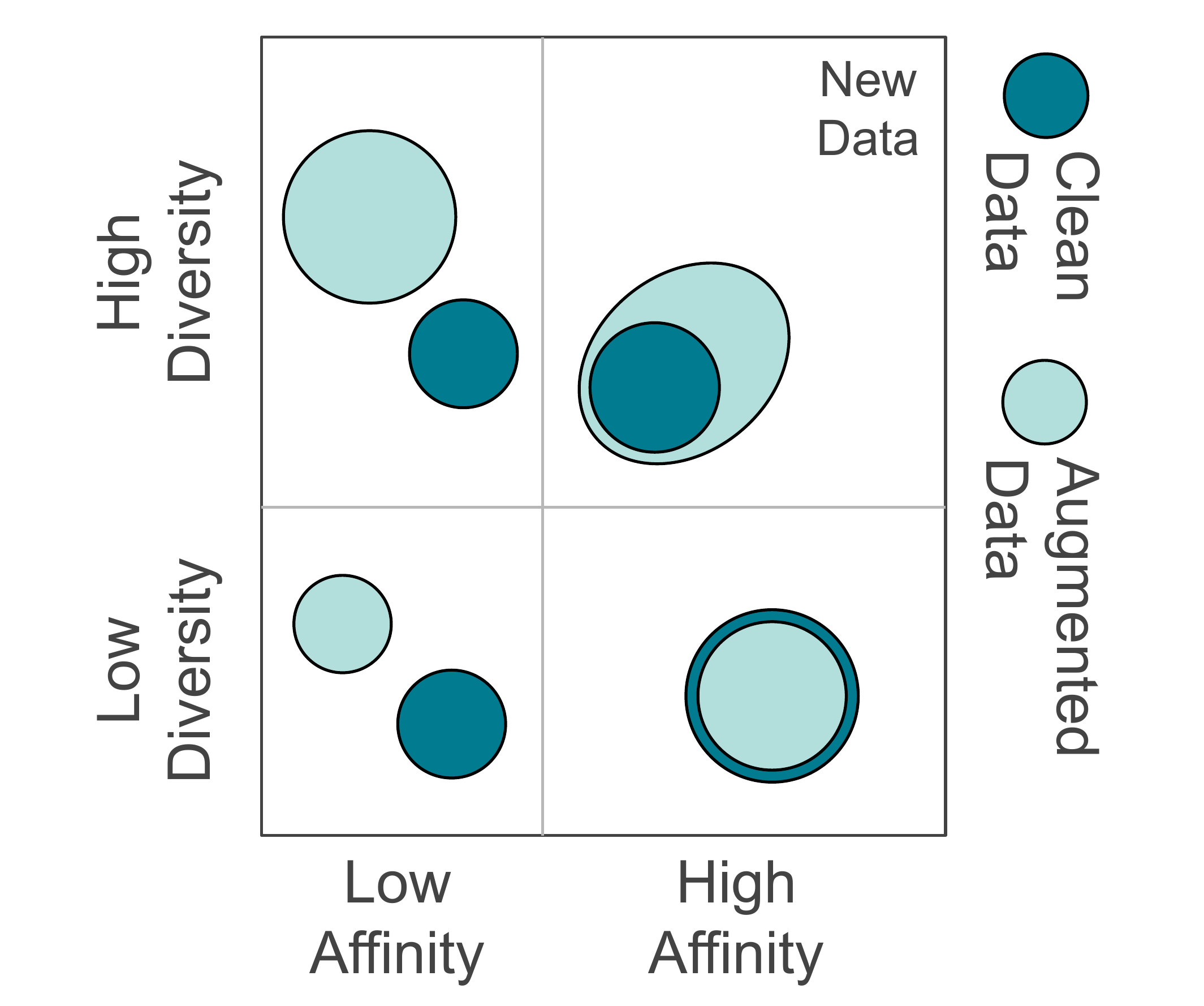}%
    }\end{subfigure}%
    \hfill%
}%
\caption{\small {\bf \indistness{} and \diversity{} parameterize the performance of a model trained with augmentation}. (a) \cifar{}: Color shows the final test accuracy. * marks the clean baseline. Each point represents a different augmentation that yields test accuracy greater than 88.7\%. (b) Representation of how clean data and augmented data are related in the space of these two metrics. Higher diversity is represented by a larger bubble while distributional similarity is depicted through the overlap of bubbles.
Test accuracy generally improves to the upper right in this space. Adding real new data to the training set is expected to be in the far upper right corner.}
\label{fig:header}
\end{center}
\vskip -0.1in
\end{figure}


\section{Introduction}
\vspace{-0.1in}%
Models that achieve state-of-the-art in image classification often use heavy data augmentation strategies.
The best techniques use various transforms applied sequentially and stochastically. Though the effectiveness of this is well-established, the mechanism through which these transformations work is not well-understood. 

Since early uses of data augmentation, it has been assumed that augmentation works because it simulates realistic samples from the true data distribution: ``[augmentation strategies are] reasonable since the transformed reference data is now extremely close to the original data. In this way, the amount of training data is effectively increased"~\citep{Bellegarda1992}. Because of this, augmentations have often been designed with the heuristic of incurring minimal distribution shift from the training data.

This rationale does not explain why unrealistic distortions such as cutout~\citep{cutout2017}, SpecAugment~\citep{park2019specaugment}, and mixup~\citep{zhang2017mixup} significantly improve generalization performance. 
Furthermore, methods do not always transfer across datasets---\aug{Cutout}, for example, is useful on \cifar{} and not on \imagenet{}~\citep{lopes2019improving}. 
Additionally, many augmentation policies heavily modify images by stochastically applying multiple transforms to a single image.
Based on this observation, some have proposed that augmentation strategies are effective because they increase the diversity of images seen by the model.

In this complex landscape, claims about diversity and distributional similarity remain unverified heuristics. Without more precise data augmentation science, finding state-of-the-art strategies requires brute force that can cost thousands of GPU hours~\citep{cubuk2018autoaugment, zhang2019adversarial}. This highlights a need to specify and measure the relationship between the original training data and the augmented dataset, as relevant to a given model's performance.

In this paper, we quantify these heuristics. Seeking to understand the mechanisms of augmentation, we focus on single transforms as a foundation. We present an extensive study of 204 different augmentations on \cifar\ and 223 on \imagenet{}, varying both broad transform families and finer transform parameters. Our contributions are:
\begin{enumerate}%
\item We introduce \indistness{} and \diversity{}: interpretable, easy-to-compute metrics for parameterizing augmentation performance. \indistness\ quantifies how much an augmentation shifts the training data distribution from that learned by a model.
\diversity{} quantifies the complexity of the augmented data with respect to the model and learning procedure. 
\item We show that performance is dependent on {\it both} metrics. In the \indistness{}-\diversity{} plane, the best augmentation strategies jointly optimize the two (see Fig~\ref{fig:header}).
\item We connect augmentation to other familiar forms of regularization, such as $\ell_{2}$ and learning rate scheduling, observing common features of the dynamics: performance can be improved and training accelerated by turning off regularization at an appropriate time. 
\item We show that performance is only improved when a transform increases the total number of unique training examples. 
The utility of these new training examples is informed by the augmentation’s \indistness{} and \diversity{}.
\end{enumerate}%

\section{Related Work}\label{Sec:related}
\vspace{-0.1in}%
Since early uses of data augmentation in training neural networks, there has been an assumption that effective transforms for data augmentation are those that produce images from an ``overlapping but different" distribution~\cite{bengio2011deep, Bellegarda1992}.
Indeed, elastic distortions as well as distortions in the scale, position, and orientation of training images have been used on \mnist~\cite{ciregan2012multi,sato2015apac,simard2003best,wan2013regularization}, while horizontal flips, random crops, and random distortions to color channels have been used on \cifar{} and \imagenet{}~\cite{krizhevsky2012imagenet,WRN2016, zoph2017learning}.
For object detection and image segmentation, one can also use object-centric cropping~\cite{liu2016ssd} or cut-and-paste new objects~\cite{dwibedi2017cut, fang2019instaboost, ngiam2019starnet}.

In contrast, researchers have also successfully used more generic transformations that are less domain-specific, such as Gaussian noise~\cite{ford2019adversarial, lopes2019improving}, input dropout~\cite{srivastava2014dropout}, erasing random patches of the training samples during training~\cite{cutout2017, park2019specaugment, zhong2017random}, and adversarial noise~\cite{szegedy2013intriguing}. Mixup~\cite{zhang2017mixup} and Sample Pairing~\cite{inoue2018data} are two augmentation methods that use convex combinations of training samples. 

It is also possible to improve generalization by combining individual transformations. For example, reinforcement learning has been used to choose more optimal combinations of data augmentation transformations~\cite{ratner2017learning, cubuk2018autoaugment}. Follow-up research has lowered the computation cost of such optimization, by using population based training~\cite{ho2019population}, density matching~\cite{lim2019fast}, adversarial policy-design that evolves throughout training~\cite{zhang2019adversarial}, or a reduced search space~\cite{cubuk2019randaugment}. Despite producing unrealistic outputs, such combinations of augmentations can be highly effective in different tasks~\cite{berthelot2019remixmatch,tan2019efficientnet, tan2019efficientdet, xie2019adversarial, xie2019unsupervised}.

Across these different examples, the role of distribution shift in training remains unclear.
\citet{lim2019fast,hataya2019faster} have found augmentation policies by minimizing the distance between the distributions of augmented data and clean data. Recent work found that after training with augmented data, fine-tuning on clean training data can be beneficial~\citep{He2019switching}, while \citet{touvron2019fixing} found it beneficial to fine-tune with a test-set resolution that aligns with the training-set resolution. 

The true input-space distribution from which a training dataset is drawn remains elusive. To better understand the effect of distribution shift on model performance, many works attempt to estimate it. Often these techniques require training secondary models, such as those based on variational methods ~\citep{NIPS2014_5423, kingma2014autoencoding, nowozin2016fgan, Blei_2017}. Others have tried to augment the training set by modelling the data distribution directly~\cite{tran2017bayesian}.
Recent work has suggested that even unrealistic distribution modelling can be beneficial~\citep{dai2017good}.

These methods try to specify the distribution separately
from the model they are trying to optimize. As a result, they are insensitive to any interaction between the model and data distribution. Instead, we are interested in a measure of how much the data shifts along directions that are most relevant to the model's performance.

\section{Methods}\label{Sec:methods}
\vspace{-0.1in}%
We performed extensive experiments with various augmentations on \cifar{} and \imagenet{}.
Experiments on \cifar{} used the WRN-28-2 model \cite{WRN2016}, trained for 78k steps with cosine learning rate decay. Results are the mean over 10 initializations and reported errors (often too small to show on figures) are the standard error on the mean. Details on the error analysis are in Sec.~\ref{sec:error}.

Experiments on \imagenet{} used the ResNet-50 model~\cite{he2016deep}, trained for 112.6k steps with a weight decay rate of 1e-4, and a learning rate of 0.2, which is decayed by 10 at epochs 30, 60, and 80.

Images were pre-processed by dividing each pixel value by 255 and normalizing by the data set statistics. Random crop was also applied on all \imagenet{} models.
These pre-processed data without further augmentation are ``clean data'' and a model trained on it is the ``clean baseline''. We followed the same implementation details as~\citet{cubuk2018autoaugment}\footnote{Available at \texttt{bit.ly/2v2FojN}}, including for most augmentation operations. Further implementation details are in Sec.~\ref{sec:training}.

For \cifar{}, test accuracy on the clean baseline is $89.7\pm0.1\%$. The validation accuracy is $89.9\pm0.2\%$. On \imagenet{}, the test accuracy is 76.06\%.

Unless specified otherwise, data augmentation was applied following standard practice: each time an image is drawn, the given augmentation is applied with a given probability. We call this mode \emph{dynamic} augmentation. Due to whatever stochasticity is in the transform itself (such as randomly selecting the location for a crop) or in the policy (such as applying a flip only with 50\% probability), the augmented image could be different each time. Thus, most of the tested augmentations increase the number of possible distinct images that can be shown during training.

We also performed select experiments using \emph{static} training. In static augmentation, the augmentation policy (one or more transforms) is applied once to the entire clean training set. Static augmentation does not change the number of unique images in the dataset.

\subsection{\indistness: a simple metric for distribution shift}
\vspace{-0.1in}%
Thus far, heuristics of distribution shift have motivated design of augmentation policies. Inspired by this focus, we introduce a simple metric to quantify how augmentation shifts data {\it with respect to the decision boundary of the clean baseline model}.

We start by noting that a trained model is often sensitive to the distribution of the training data. That is, model performance varies greatly between new samples from the true data distribution and samples from a shifted distribution. 

Importantly, the model’s sensitivity to distribution shift is not purely a function of the input data distribution, since training dynamics and the model’s implicit biases affect performance. Because the goal of augmentation is improving model performance, measuring shifts with respect to the distribution captured by the model is more meaningful than measuring shifts in the distribution of the input data alone. 

We thus define \indistness\ to be the difference between the validation accuracy of a model trained on clean data and tested on clean data, and the accuracy of the same model tested on an augmented validation set. Here, the augmentation is applied to the validation dataset in one pass, as a static augmentation. More formally we define:
\begin{definition}%
Let ${D}_{\textrm{train}}$ and ${D}_{\textrm{val}}$ be training and validation datasets drawn IID from the same \emph{clean} data distribution, and let ${D}'_{\textrm{val}}$ be derived from ${D}_{\textrm{val}}$ by applying a stochastic augmentation strategy, $a$, once to each image in ${D}_{\textrm{val}}$, ${D}'_{\textrm{val}}=\{(a(x), y)\ :\forall\ (x,y)\in{D}_{\textrm{val}}\}$. Further let $m$ be a model trained on ${D}_{\textrm{train}}$ and $\mathcal{A}(m,{D})$ denote the model's accuracy when evaluated on dataset ${D}$. The \indistness, $\mathcal{T}[a;m;{D}_\textrm{val}]$, is given by
\begin{align}
    \mathcal{T}[a; m; {D}_\textrm{val}]=\mathcal{A}(m,{D}'_\textrm{val}) - \mathcal{A}(m,{D}_\textrm{val})\,.
\end{align}
\end{definition}%
With this definition, \indistness\ of zero represents no shift and a negative number suggests that the augmented data is out-of-distribution for the model.

\begin{figure}[ht]
\vspace{-0.2in}
\vskip 0.1in 
\begin{center}
\begin{minipage}[l]{0.5\textwidth}
        \begin{subfigure}[\indistness]{%
            \includegraphics[width=0.5\columnwidth]{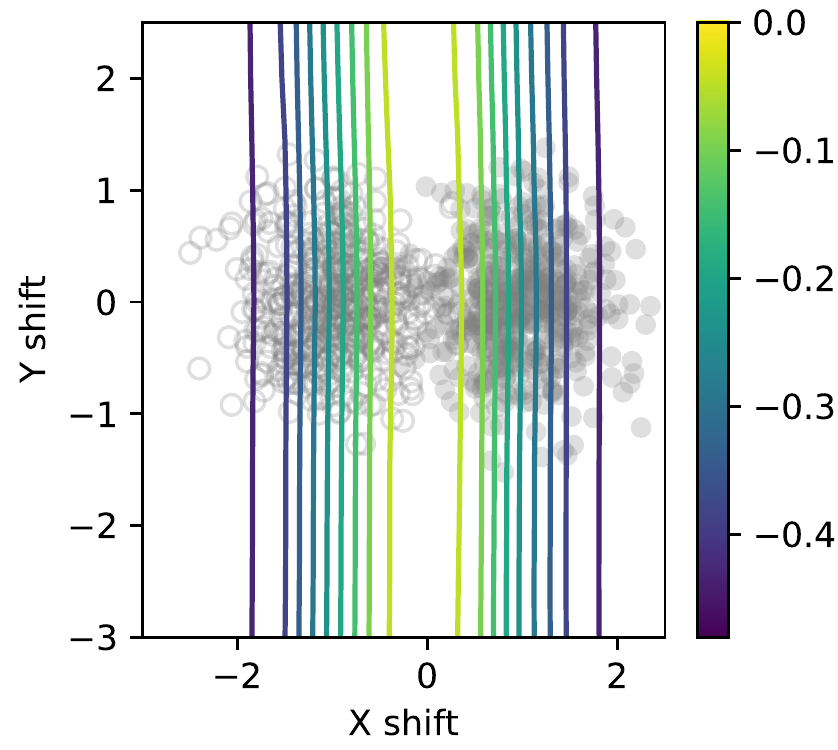}%
        }\end{subfigure}%
        \hfill%
        \begin{subfigure}[$D_{\textrm{KL}}$]{%
            \includegraphics[width=0.5\columnwidth]{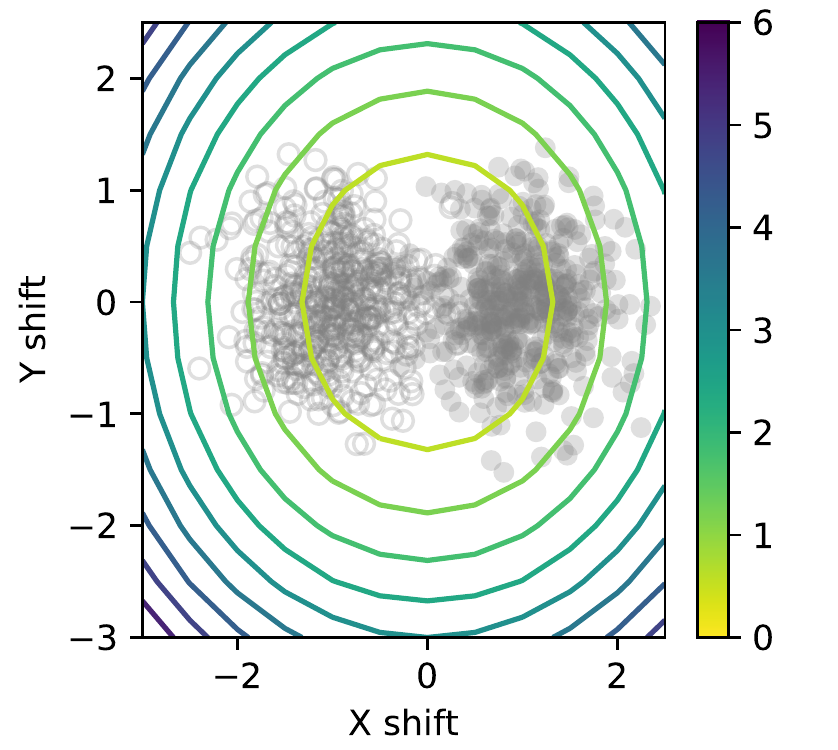}%
        }\end{subfigure}%
\end{minipage}\hfill
\raisebox{0.2in}{
\begin{minipage}[c]{0.48\textwidth}
    \caption{\small {\bf \indistness{} is a model-sensitive measure of distribution shift}. Contours indicate lines of equal (a) \indistness{}, or (b) KL Divergence between the joint distribution of the original data and targets and the shifted data. The two axes indicate the actual shifts that define the augmentation. \indistness{} captures model-dependent features, such as the decision boundary.}
    \label{fig:toy_metrics}
\end{minipage}
}
\end{center}
\vspace{-0.2in}
\end{figure}

In Fig.~\ref{fig:toy_metrics} we illustrate \indistness{} with a two-class classification task on a mixture of two Gaussians. Augmentation in this example comprises shift of the means of the Gaussians of the validation data compared to those used for training. Under this shift, we calculate both \indistness{} and KL divergence of the shifted data with respect to the original data. \indistness{} changes only when the shift in the data is with respect to the model's decision boundary, whereas the KL divergence changes even when data is shifted in the direction that is irrelevant to the classification task. In this way, \indistness{} captures what is relevant to a model: shifts that impact predictions.

This same metric has been used as a measure of a model's robustness to image corruptions that do not change images' semantic content \citep{azulay2018deep, dodge2017study,ford2019adversarial,hendrycks2018benchmarking,  rosenfeld2018elephant, yin2019afourier}. Here we, turn this around and use it to quantify the shift of augmented data compared to clean data.

\indistness{} has the following advantages as a metric:%
\begin{enumerate}%
\item It is easy to measure. It requires only clean training of the model in question. %
\item It is independent of any confounding interaction between the data augmentation and the training process, since augmentation is only used on the validation set and applied statically.%
\item It is a measure of distance sensitive to properties of both the data distribution \emph{and} the model.%
\end{enumerate}%

We gain confidence in this metric by comparing it to other potential model-dependant measures of distribution shift. We consider the mean log likelihood of augmented test images\citep{grathwohl2019energyclassifier}, and the Watanabe–Akaike information criterion (WAIC) \citep{WAIC}. These other metrics have high correlation with \indistness{}. Details can be found in Sec.~\ref{sec:supplsewaic}.

\subsection{Diversity: A measure of augmentation complexity}
\vspace{-0.1in}%
Inspired by the observation that multi-factor augmentation policies such as \aug{FlipLR}+\aug{Crop}+\aug{Cutout} and \aug{RandAugment}\citep{cubuk2019randaugment} greatly improve performance, we propose another axis on which to view augmentation policies, which we dub \emph{\diversity}. This measure is intended to quantify the intuition that augmentations prevent models from over-fitting by increasing the number of samples in the training set; the importance of this is shown in Sec.~\ref{sec:fixedaug}.

Based on the intuition that more diverse data should be more difficult for a model to fit, we propose a model-based measure. The \diversity{} metric in this paper is the final training loss of a model trained with a given augmentation:
\begin{definition}\label{defn:diversity}
Let $a$ be an augmentation  and $D_{\textrm{train}}'$ be the augmented training data resulting from applying the augmentation, $a$, stochastically. Further, let $L_{\textrm{train}}$ be the training loss for a model, $m$, trained on $D_{\textrm{train}}'$. We define the \diversity, $\mathcal{D}[a;m;D_{\textrm{train}}]$, as
\begin{align}
    \mathcal{D}[a;m;D_{\textrm{train}}]&:=\mathbb{E}_{ D_{\textrm{train}}'}\left[L_{\textrm{train}}\right]\,.
\end{align}
\end{definition}

Though determining the training loss requires the same amount of work as determining final test accuracy, here we focus on this metric as a tool for understanding. 

As with \indistness{}, this definition of \diversity{} has the advantage that it can capture model-dependent elements, i.e. it is informed by the class of priors implicit in choosing a model and optimization scheme as well as by the stopping criterion used in training.

Another potential diversity measure is the entropy of the transformed data, $\mathcal{D}_{\textrm{Ent}}$. This is inspired by the intuition that augmentations with more degrees of freedom perform better. For discrete transformations, we consider the conditional entropy of the augmented data.
\begin{align}
    \mathcal{D}_{\textrm{Ent}}\,:=\,H(X'|X)\,=\,-\mathbb{E}_{X}\left[\Sigma_ {x'} p(x'|X)\log(p(x'|X))\right]\,.\nonumber
\end{align}
Here $x\in X$ is a clean training image and $x' \in X'$ is an augmented image. This measure has the property that it can be evaluated without any training or reference to model architecture. However, the appropriate entropy for continuously-varying transforms is less straightforward. 

A third proxy for \diversity{} is the training time needed for a model to reach a given training accuracy threshold. In Sec.~\ref{sec:divmetrics}, we show that these three metrics correlate well with each other.

In the remaining sections we describe how the complementary metrics of \diversity\ and \indistness\ can be used to characterize and understand augmentation performance.

 
\section{Results}
\vspace{-0.1in}%
\subsection{Augmentation performance is determined by both \indistness{} and \diversity{}}
\vspace{-0.1in}%
Despite the original inspiration to mimic realistic transformations and minimize distribution shift, many state-of-the-art augmentations yield unrealistic images.
This suggests that distribution shift alone does not fully describe or predict augmentation performance.

\begin{figure}[ht]
    \centering%
    \begin{subfigure}[T: \cifar{}. B: \imagenet{}]{%
        \begin{minipage}[b]{0.3\columnwidth}%
            \includegraphics[height=0.72in]{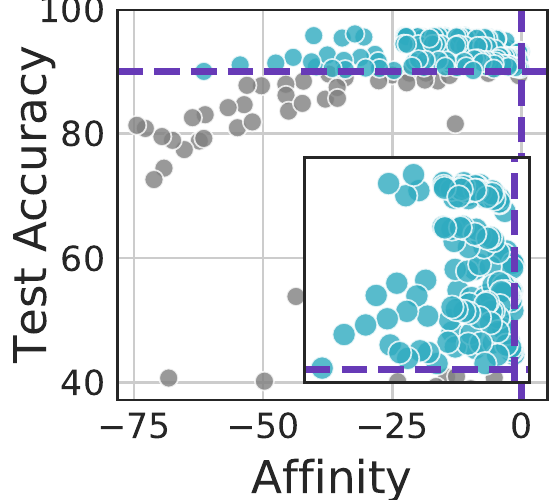}%
            \includegraphics[height=0.72in, clip, trim={0.5cm 0 0 0}]{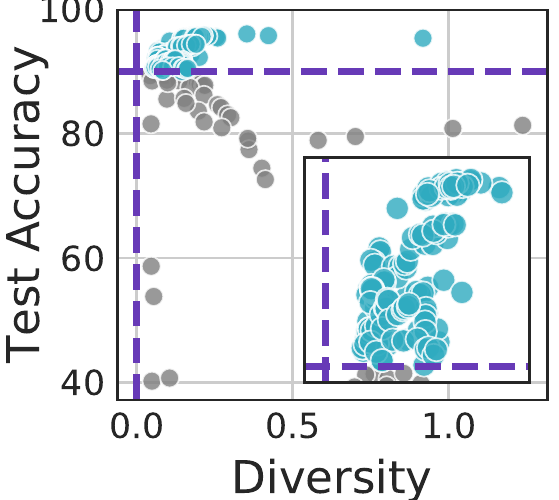}
            \vskip 0pt\vspace{-0.1in}%
            \includegraphics[height=0.72in]{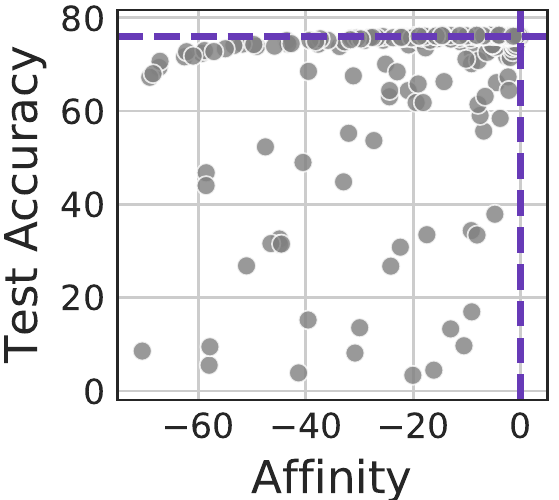}%
            \includegraphics[height=0.72in, clip, trim={0.5cm 0 0 0}]{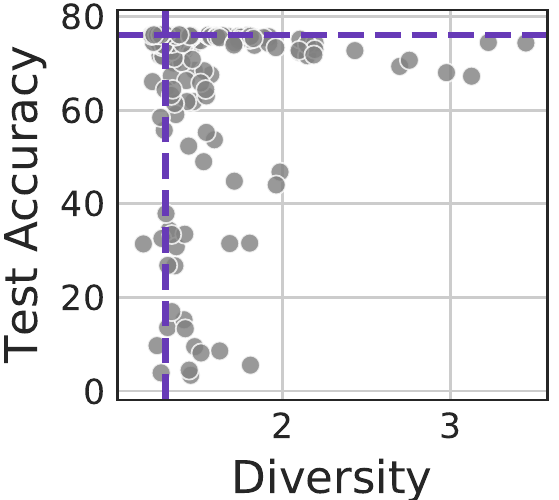}%
            \vspace{0.1in}
         \end{minipage}%
         \label{huge-a}
    }\end{subfigure}%
    \hfill%
    \begin{subfigure}[\cifar{}]{%
         \includegraphics[width=0.34\columnwidth]{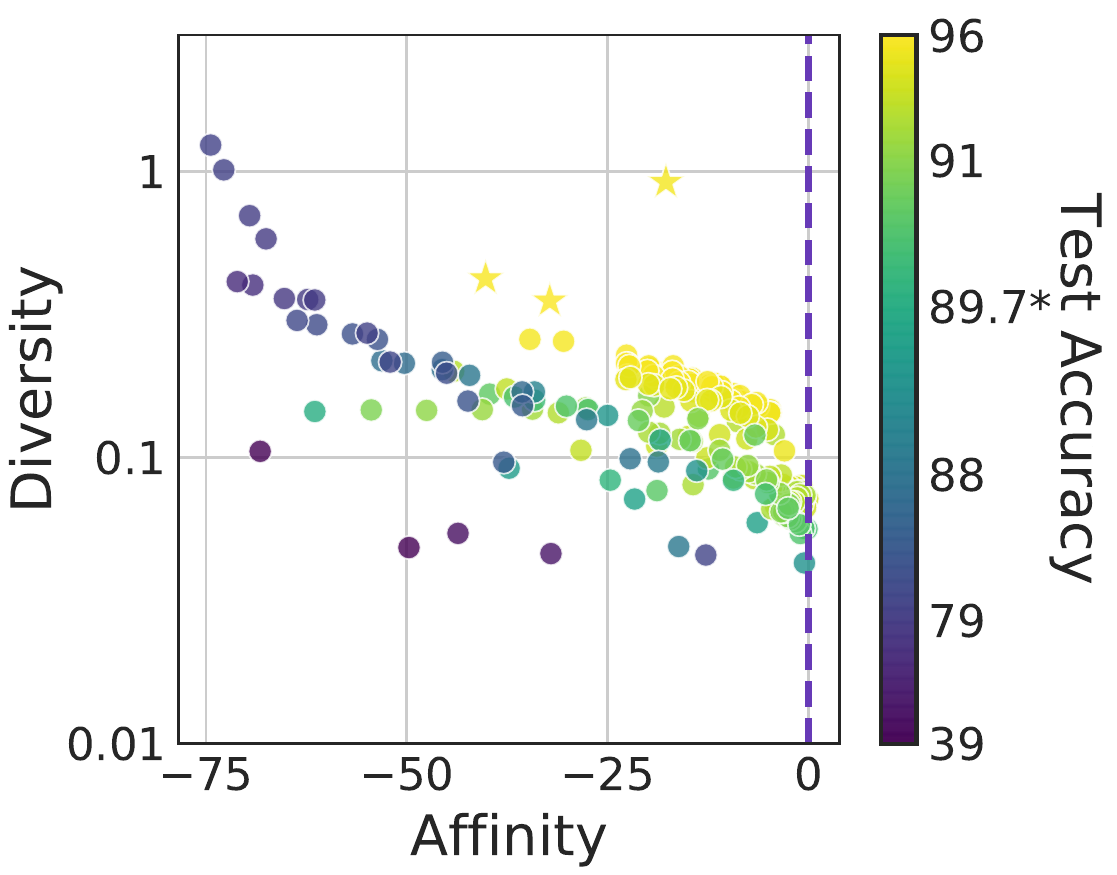}%
         \label{huge-b}
    }\end{subfigure}%
    \hfill%
    \begin{subfigure}[\imagenet{}]{%
         \includegraphics[width=0.34\columnwidth]{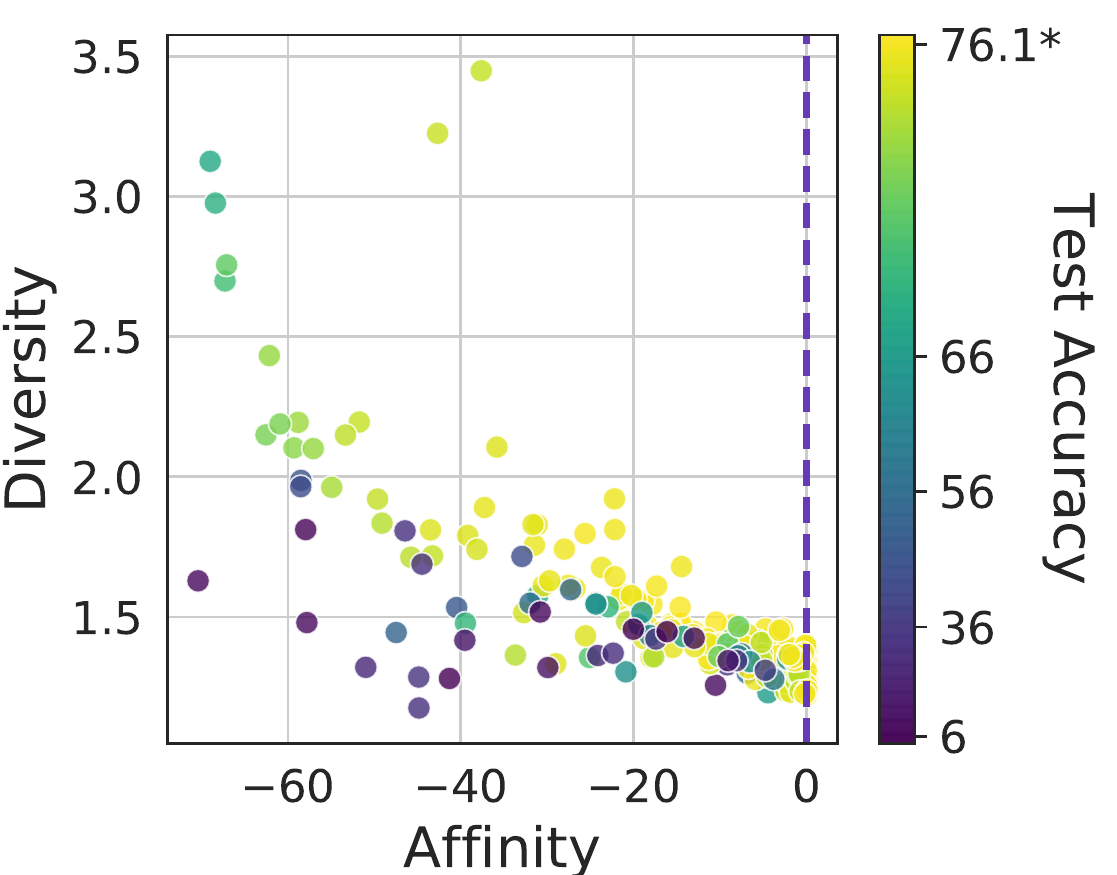}%
         \label{huge-c}
    }\end{subfigure}%
    \caption{\small {\bf Augmentation performance is determined by both \indistness{} and \diversity{}}. (a) Test accuracy plotted against each of \indistness{} and \diversity{} for the two datasets, showing that neither metric alone predicts performance. In the \cifar{} plots (top), blue highlights (also in inset) are the augmentations that increase test accuracy above the clean baseline. Dashed lines indicate the clean baseline. (b) and (c) show test accuracy on the color scale in the plane of \indistness{} and \diversity{}. The three star markers in (b) are (left to right) \aug{RandAugment}, \aug{AutoAugment}, and \aug{mixup}. The * on the color bar indicates the clean baseline case. For fixed values of \indistness{}, test accuracy generally increases with higher values of \diversity{}. For fixed values of \diversity{}, test accuracy generally increases with higher values of \indistness{}. Note that the gains observed on \imagenet{} are expected to be small, in line with previous work on single-transformation policies~\cite{yin2019afourier}.}
    \label{fig:mainhugefig}
\end{figure}

Figure~\ref{huge-a} (left) measures \indistness\ across 204 different augmentations for \cifar\ and 223 for \imagenet\ respectively. We find that for the most important augmentations---those that help performance---\indistness\ is a poor predictor of accuracy. Furthermore, we find many successful augmentations with low \indistness{}. For example, \aug{Rotate(fixed, 45deg, 50\%)}, \aug{Cutout(16)}, and combinations of \aug{FlipLR}, \aug{Crop(32)}, and \aug{Cutout(16)} all have \indistness{}$<-15\%$ and test accuracy$>2\%$ above clean baseline on \cifar{}. Augmentation details are in Sec.~\ref{sec:augdetails}.

As \indistness{} does not fully characterize the performance of an augmentation, we seek another metric. To assess the importance of an augmentation's complexity, we measure \diversity\ across the same set of augmentations. We find that \diversity{} is complementary in explaining how augmentations can increase test performance. As shown in Fig.~\ref{huge-b} and \subref{huge-c}, \indistness\ and \diversity\ together provide a much clearer parameterization of an augmentation policy's benefit to performance. For a fixed level of \diversity, augmentations with higher \indistness\ are consistently better. Similarly, for a fixed \indistness, it is generally better to have higher \diversity{}.

\begin{figure}[ht]
\vskip 0.1in 
\vspace{-0.2in}
\begin{center}
\centerline{%
    \raisebox{15 pt}{
        \begin{subfigure}{%
            \includegraphics[width=0.23\columnwidth]{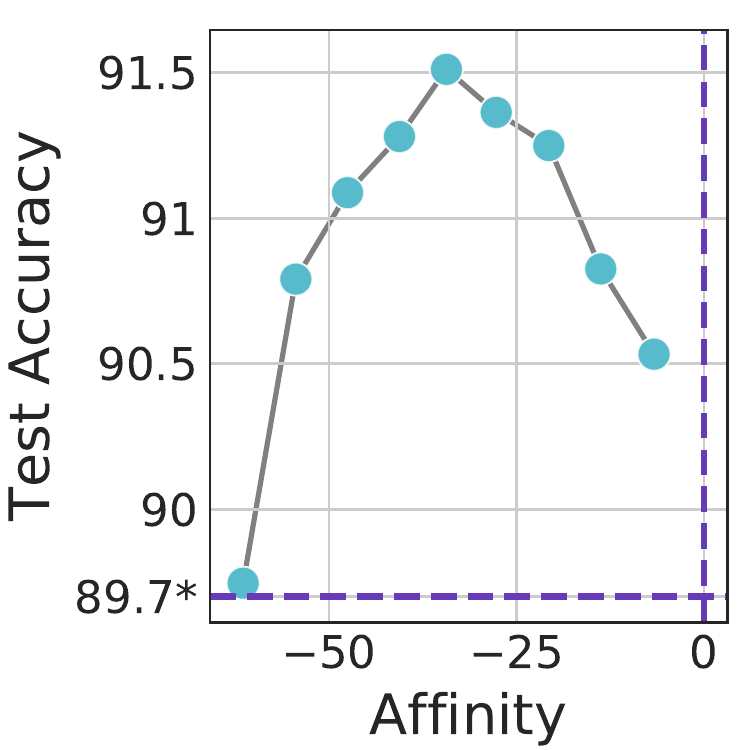}%
        }\end{subfigure}%
        \begin{subfigure}{%
            \includegraphics[width=0.23\columnwidth]{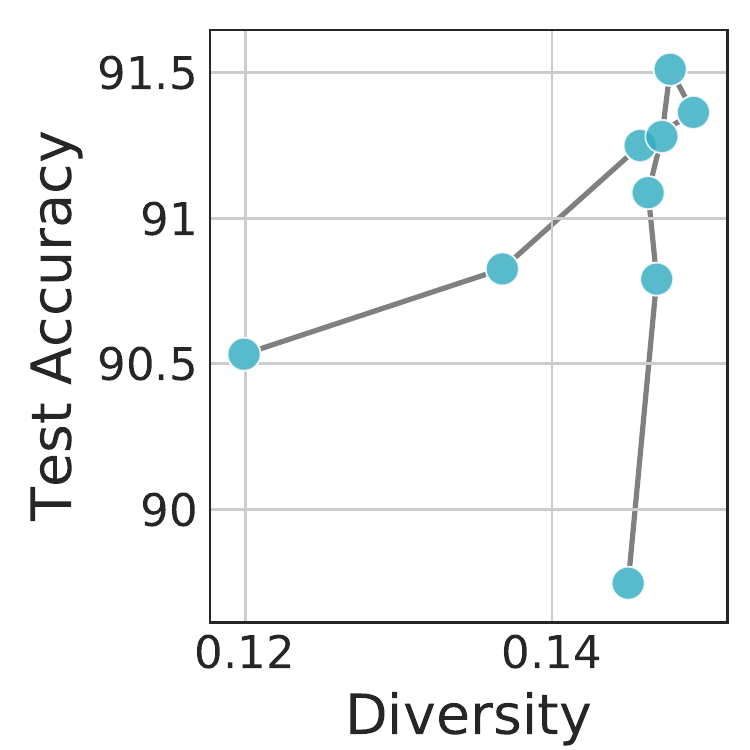}%
        }\end{subfigure}%
    }
        \begin{subfigure}{%
            \includegraphics[width=0.35\columnwidth]{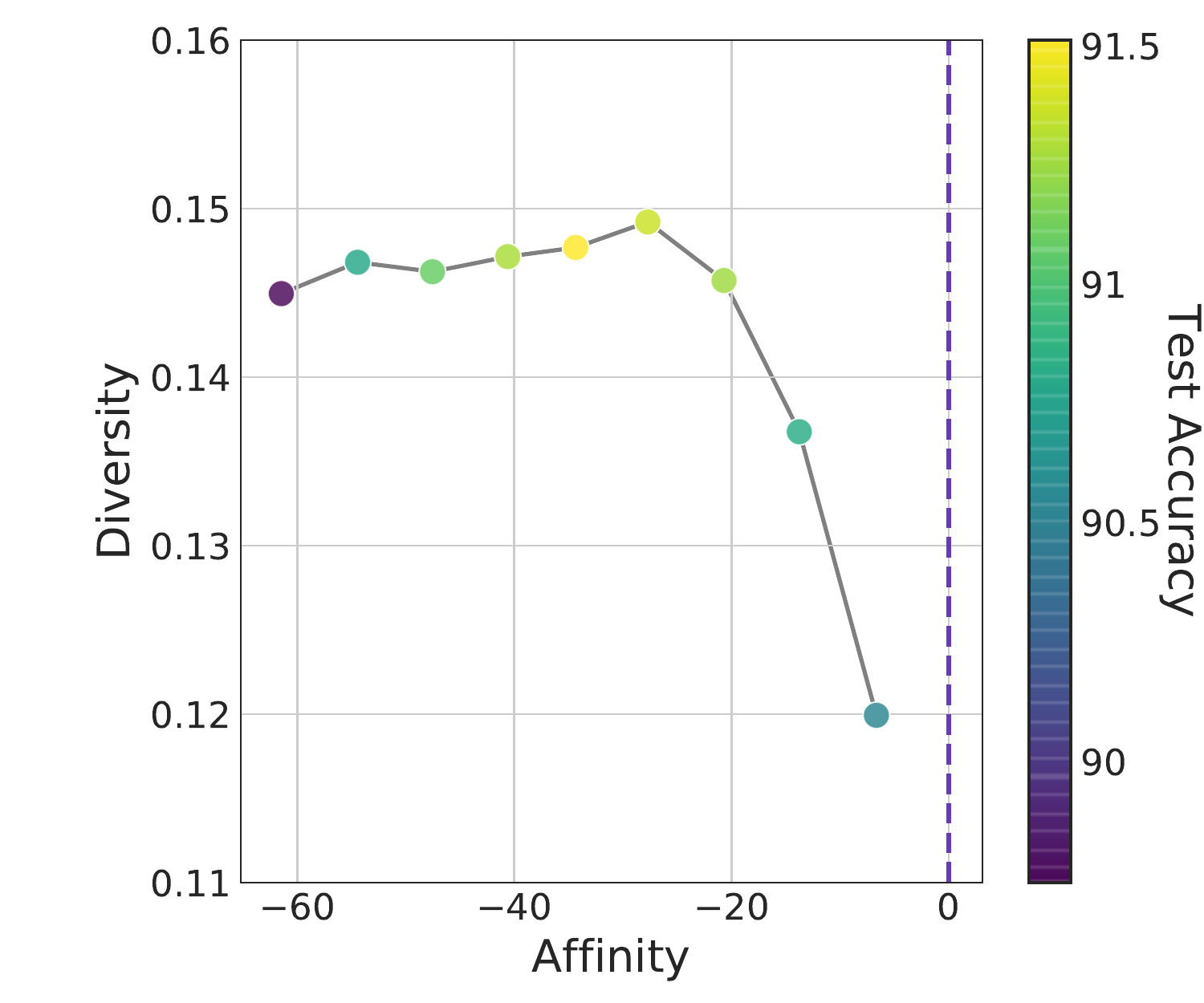}
        }\end{subfigure}%
}%
\vspace{-0.1in}%
\caption{\small {\bf Test accuracy varies differently than either \indistness{} or \diversity}. Here, the probability of \aug{Rotate(fixed, 60deg)} on \cifar{} is varied from 10\% to 90\%. Left: as probability increases, \indistness{} decreases linearly while the accuracy changes non-monotonically. Center: accuracy and \diversity\ vary differently from each other as probability is changed. Right: test accuracy is maximized at intermediate values.}
\label{fig:rotate_case_study}
\end{center}
\vskip -0.2in
\end{figure}

A simple case study is presented in Fig.~\ref{fig:rotate_case_study}. The probability of the transform \aug{Rotate(fixed, 60deg)} is varied. The accuracy and \indistness\ are not monotonically related, with the peak accuracy falling at an intermediate value of \indistness. Similarly, accuracy is correlated with \diversity{} for low probability transformations, but does not track for higher probabilities. The optimal probability for \aug{Rotate(fixed, 60deg)} lies at an intermediate value of \indistness{} and \diversity{}.

To situate the tested augmentations---mostly single transforms---within the context of the state-of-the-art, we tested three high-performance augmentations from literature: \aug{mixup} ~\citep{zhang2017mixup}, \aug{AutoAugment} \citep{cubuk2018autoaugment}, and \aug{RandAugment}~\citep{cubuk2019randaugment}. These are highlighted with a star marker in Fig.~\ref{huge-b}.

More than either of the metrics alone, \indistness{} and \diversity{} together provide a useful parameterization of an augmentation's performance. 
We now turn to investigating the utility of this tool for explaining other observed phenomena of data augmentations.

\subsection{Turning augmentations off may adjust \indistness{}, \diversity{}, and performance}
\vspace{-0.1in}%

The term ``regularizer'' is ill-defined in the literature, often referring to any technique used to reduce generalization error without necessarily reducing training error~\citep{Goodfellow-et-al-2016}. With this definition, it is widely acknowledged that commonly-used augmentations act as regularizers~\citep{hernandez2018further,zhang2016understanding,dao2019kernel}. 
Though this is a broad definition, we notice another commonality across seemingly different kinds of regularizers:
various regularization techniques yield boosts in performance (or at least no degradation) if the regularization is \textit{turned off} at the right time during training. For instance:%
\begin{enumerate}%
    \item Decaying a large learning rate on an appropriate schedule can be better than maintaining a large learning rate throughout training~\citep{WRN2016}.%
    \item Turning off $\ell_{2}$ regularization at the right time in training does not hurt performance~\citep{golatkar2019time}.%
    \item Relaxing architectural constraints mid-training can boost final performance~\citep{dascoli2019CNNtoFCN}.%
    \item Turning augmentations off and fine-tuning on clean data can improve final test accuracy~\citep{He2019switching}.%
\end{enumerate}%
To further study augmentation as a regularizer, we compare the constant augmentation case (with the same augmentation throughout) to the case where the augmentation is turned off partway through training and training is completed with clean data. For each transform, we test over a range of switch-off points and select the one that yields the best final validation or test accuracy on \cifar{} and \imagenet{} respectively. The Switch-off Lift is the resulting increase in final test accuracy, compared to training with augmented data the entire time. 

\begin{figure}[ht!]
\vskip 0.1in 
\begin{center}
    \begin{subfigure}[Slingshot effect on \cifar{}]{
        \includegraphics[width=0.45\columnwidth]{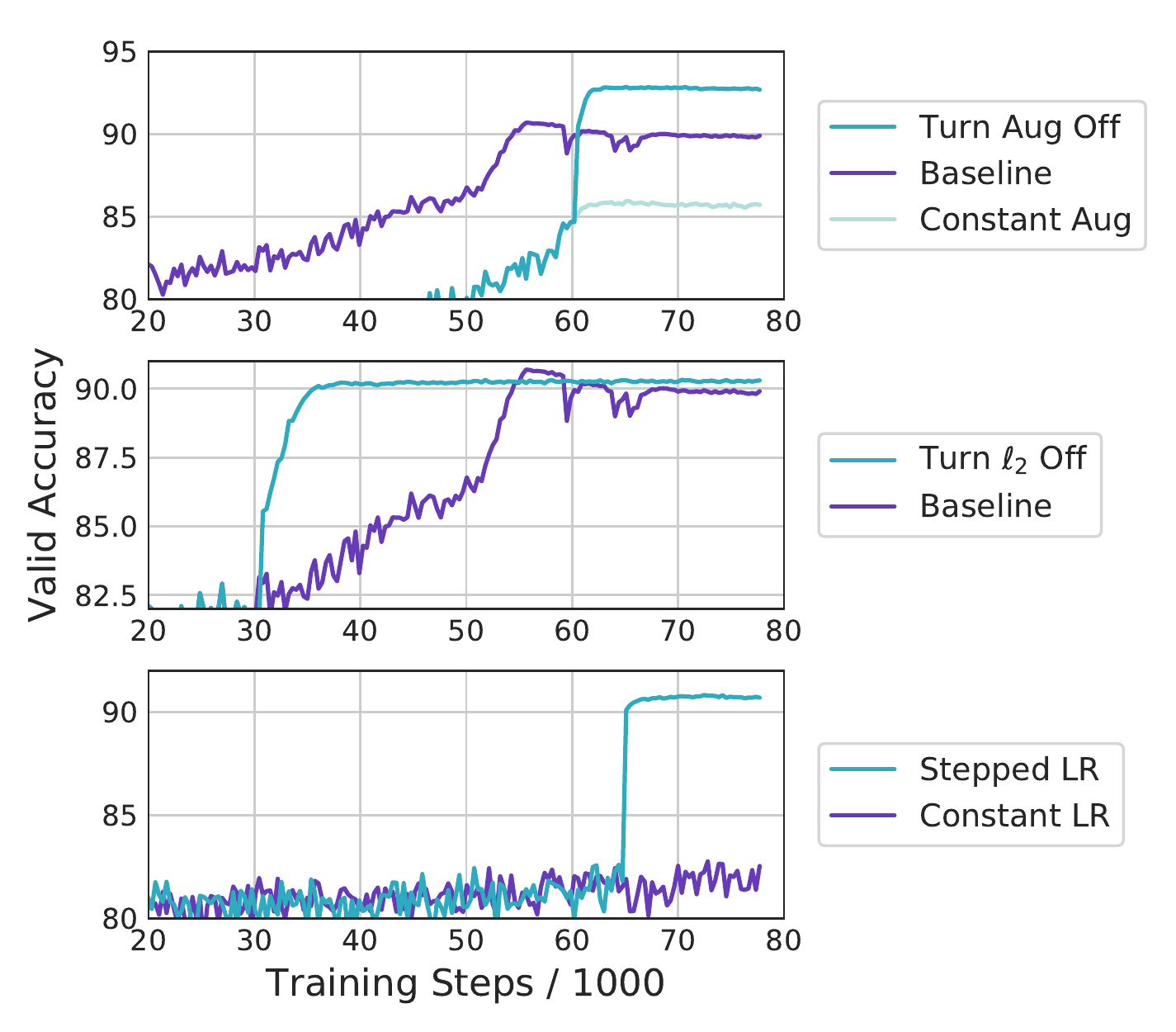}%
        \label{slingshot-a}
    }\end{subfigure}
    \begin{minipage}[b]{0.45\columnwidth}
        \begin{subfigure}[Switch-off Lift on \cifar{}]{
            \includegraphics[width=\columnwidth]{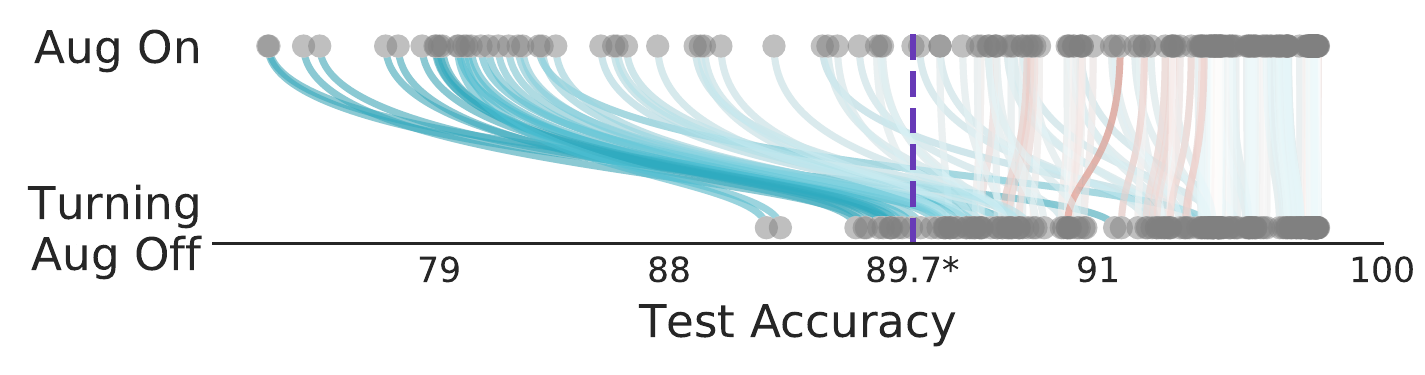}%
            \label{slingshot-b}
        }\end{subfigure}
        \hfill %
        \begin{subfigure}[\cifar{} (left) \imagenet{} (right)]{
            \includegraphics[width=0.495\columnwidth]{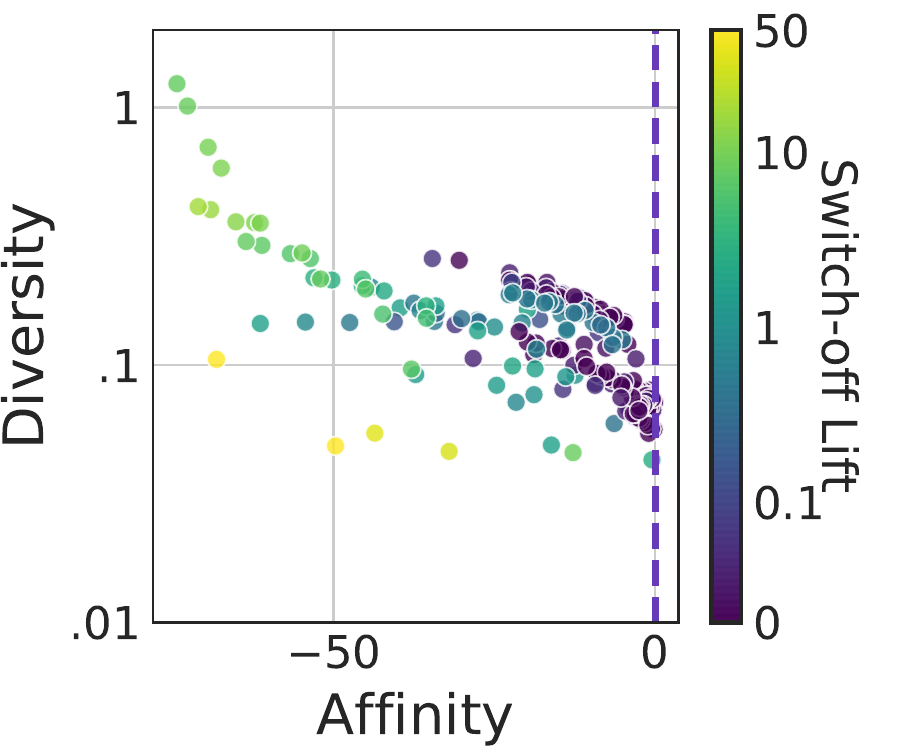}%
            \includegraphics[width=0.495\columnwidth]{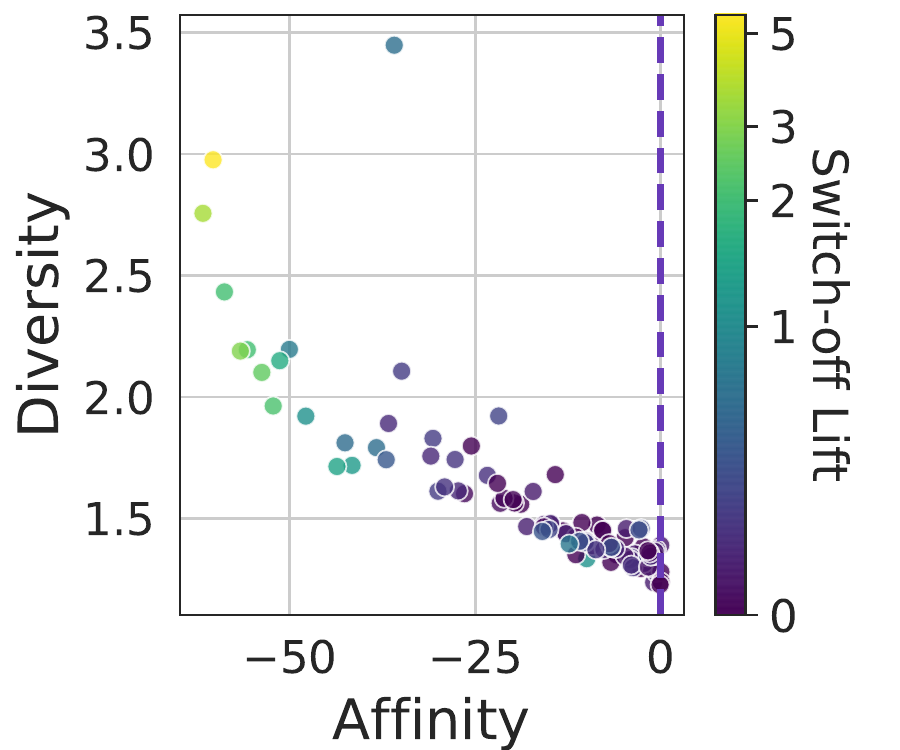}%
            \label{subfig:switching_DnI}
        }\end{subfigure}%
    \end{minipage}
\caption{\small {\bf \subref{slingshot-a} Switching off regularizers yields a performance boost}: Three examples of how turning off a regularizer increases the validation accuracy. This slingshot effect can speed up training and improve the best validation accuracy. Top: training with no augmentation (clean baseline), compared to constant augmentation, and augmentation that is turned off at 55k steps. Here, the augmentation is \aug{Rotate(fixed, 20deg,100\%)}. Middle: Baseline with constant $\ell_{2}$. This is compared to turning off $\ell_{2}$ regularization part way through training. Bottom: Constant learning rate of 0.1 compared to training where the learning rate is decayed in one step by a factor of 10.
{\bf \subref{slingshot-b} Bad augmentations can become helpful if switched off}: Colored lines connect the test accuracy with augmentation applied throughout training (top) to the test accuracy with switching mid-training. Color indicates the amount of Switch-off Lift; blue is positive and orange is negative.
{\bf \subref{subfig:switching_DnI} Switch-off Lift varies with \indistness{} and \diversity{}}. Where Switch-off Lift is negative, it is mapped to 0 on the color scale.}
\label{fig:slingshot}
\end{center}
\vskip -0.1in
\vspace{-0.15in}
\end{figure}

For some poor-performing augmentations, this gain can actually bring the final test accuracy above the baseline, as shown in Fig.~\ref{slingshot-b}. We additionally observe (Fig.~\ref{slingshot-a}) that this test accuracy improvement can happen quite rapidly for both augmentations and for the other regularizers tested. This suggests an opportunity to accelerate training without hurting performance by appropriately switching off regularization. We call this a \emph{slingshot} effect.

Interestingly, we find the best time for turning off an augmentation is not always close to the end of training, contrary to what is shown in \citet{He2019switching}. For example, without switching, \aug{FlipUD(100\%)} decreases test accuracy by almost 50\% compared to clean baseline. When the augmentation is used for only the first third of training, final test accuracy is above the baseline.

\citet{He2019switching} hypothesized that the gain from turning augmentation off is due to recovery from a distribution shift. Indeed, for many detrimental transformations, the test accuracy gained by turning off the augmentation merely recovers the clean baseline performance. However, in Fig.~\ref{subfig:switching_DnI}, we see that for a given value of \indistness{}, Switch-off Lift can vary. This result suggests that the Switch-off Lift is derived from more than simply correction of a distribution shift.


A few of the tested augmentations, such as \aug{FlipLR(100\%)}, are fully deterministic. Thus, each time an image is drawn in training, it is augmented the same way. When such an augmentation is turned off partway through training, the model then sees images---the clean ones---that are now new. Indeed, when \aug{FlipLR(100\%)} is switched off at the right time, its final test accuracy exceeds that of \aug{FlipLR(50\%)} without switching. In this way, switching augmentation off may adjust for not only low \indistness{} but also low \diversity{}. 


\subsection{Increased effective training set size is crucial for data augmentation}\label{sec:fixedaug}
\vspace{-0.1in}%

Most augmentations we tested and those used in practice have inherent stochasticity and thus may alter a given training image differently each time the image is drawn. In the typical \emph{dynamic} training mode, these augmentations increase the number of unique inputs seen across training epochs.

To further study how augmentations act as regularizers, we seek to discriminate this increase in effective dataset size from other effects. We train models with \emph{static} augmentation, as described in Sec.~\ref{Sec:methods}. This altered training set is used without further modification during training so that the number of unique training inputs is the same between the augmented and the clean training settings.

For almost all tested augmentations, using static augmentation yields lower test accuracy than the clean baseline. Where static augmentation shows a gain (versions of \aug{crop}), the difference is less than the standard error on the mean. As in the dynamic case, poorer performance in the static case is for transforms that have lower \indistness{} and lower \diversity{}. 

\begin{figure}[h]
\vspace{-0.1in}
\hspace{-0.1in}
\begin{minipage}[l]{0.6\textwidth}
    \hspace{-0.1in}\includegraphics[width=0.5\columnwidth]{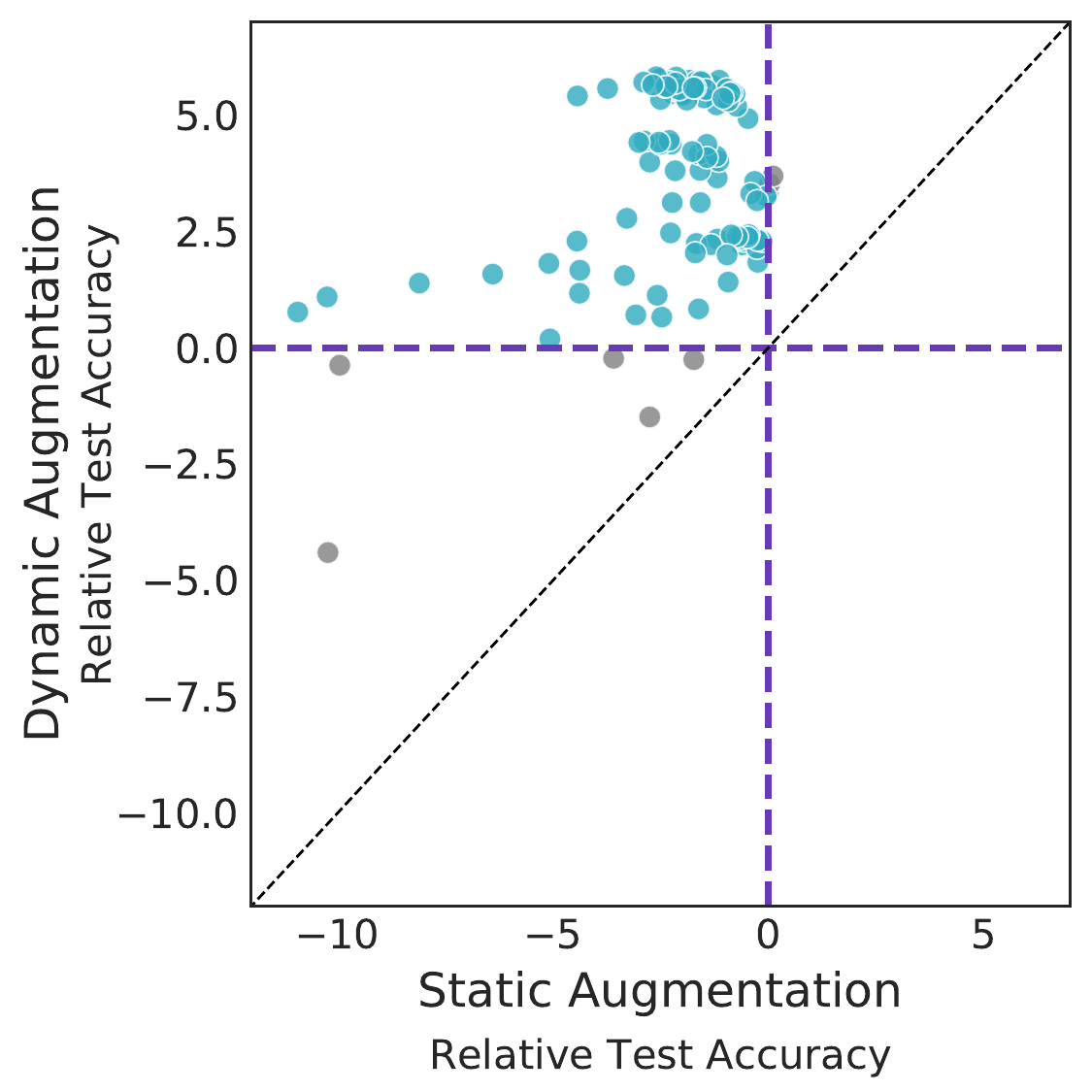}%
    \includegraphics[width=0.5\columnwidth]{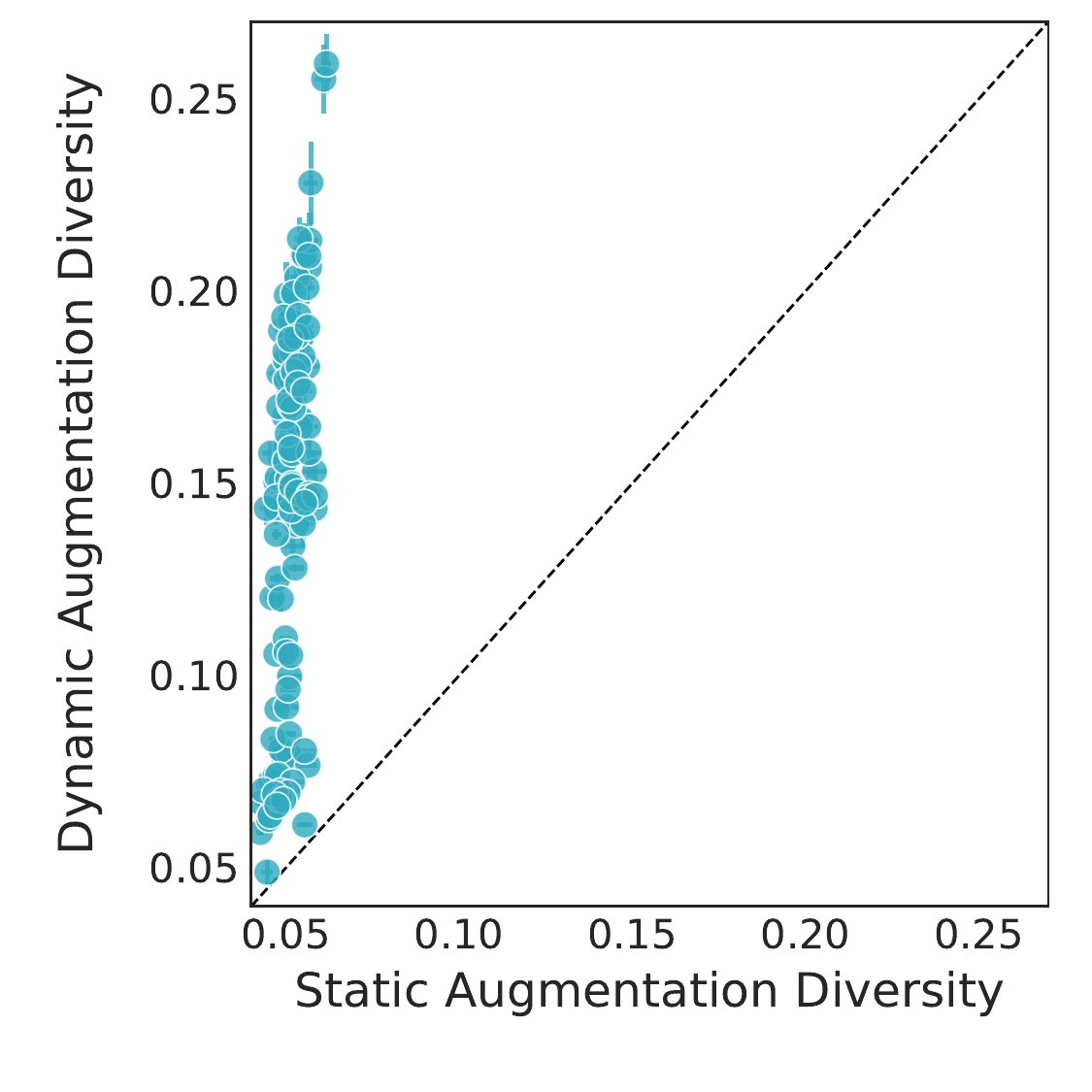}
\end{minipage}
\raisebox{0.15in}{
    \begin{minipage}[l]{0.4\textwidth}
        \caption{\small {\bf Static augmentations decrease diversity and performance}. \cifar{}, static augmentation performance is less than the clean baseline, $(0,0)$, and less than the dynamic augmentation case. Augmentations with no stochasticity are excluded because they are trivially equal on the two axes (left). \diversity{} in the static case is less than in the dynamic case.(right) Diagonal line indicates where static and dynamic cases would be equal. }\label{fig:fixedaug}
    \end{minipage}}
\vspace{-0.1in}
\end{figure}

Static augmentations also always perform worse than their non-deterministic, dynamic counterparts, as shown in Fig.~\ref{fig:fixedaug}. This may be because the \diversity{} of a static augmentation is always less than the dynamic case (see also Sec.~\ref{sec:divmetrics}). The decrease in \diversity{} in the static case suggests a connection between \diversity{} and the number of training examples. 

Together, these results point to the following conclusion: \textit{Increased effective training set size is crucial to the performance benefit of data augmentation. An augmentation's \indistness{} and \diversity{} inform how useful the additional training examples are.}
 

\section{Discussion}%
\vspace{-0.15in}%
In this work, we focused on single transforms in an attempt to understand the essential parts of augmentation in a controlled context. This builds a foundation for using these metrics to quantify and design more complex and powerful combinations of augmentations. 

Though earlier work has often explicitly focused on just one of these metrics, chosen priors have implicitly ensured reasonable values for both. One way to achieve \diversity{} is to use combinations of many single augmentations, as in AutoAugment \citep{cubuk2018autoaugment}. Because transforms and hyperparameters in~\citet{cubuk2018autoaugment} were chosen by optimizing performance on proxy tasks, the optimal policies include high and low \indistness{} transforms. 
Fast AutoAugment~\citep{lim2019fast}, CTAugment~\citep{berthelot2019remixmatch, sohn2020fixmatch}, and differentiable RandAugment~\citep{cubuk2019randaugment} all aim to increase \indistness{} by what \citet{lim2019fast} called ``density-matching''. However these methods use the search space of AutoAugment and thus inherit its \diversity{}. 

On the other hand, Adversarial AutoAugment~\cite{zhang2019adversarial} focused on increasing \diversity{} by optimizing policies to increase the training loss. While this method did not explicitly aim to increase \indistness{}, it also used transforms and hyperparameters from the AutoAugment search space. Without such a prior, which includes useful \indistness{}, the goal of maximizing training loss with no other constraints would lead to data augmentation policies that erase all the information from the images. 

Our results motivate casting an even wider net when searching for augmentation strategies. Firstly, our work suggests that explicitly optimizing along axes of both \indistness{} and \diversity{} yields better performance. Furthermore, we have seen that poor-performing augmentations can actually be helpful if turned off during training (Fig.~\ref{fig:slingshot}). With inclusion of scheduling in augmentation optimization, we expect there are opportunities for including a different set of augmentations in an ideal policy. \citet{ho2019population} observes trends in how probability and magnitude of various transforms change during training for an optimized augmentation schedule. We suggest that with further study, \diversity{} and \indistness{} can provide priors for optimization of augmentation schedules. 
\newpage
\section{Conclusion}
\vspace{-0.1in}%
We attempted to quantify common intuition that more in-distribution and more diverse augmentation policies perform well. To this end, we introduced two easy-to-compute metrics, \indistness\ and \diversity{}, intended to measure to what extent a given augmentation is in-distribution and how complex the augmentation is to learn. Because they are model-dependent, these metrics capture the data shifts that affect model performance.

With these tools, we have conducted a study over a large class of augmentations for \cifar\ and \imagenet\ and found that neither feature alone is a perfect predictor of performance.
Rather, we presented evidence that \diversity\ and \indistness\ play dual roles in determining augmentation quality. Optimizing for either metric separately is sub-optimal and the best augmentations balance the two.

Additionally, we found that an increased number of training examples, connected to \diversity{}, was a necessary ingredient of beneficial augmentation.

Finally, we found that augmentations share an important feature with other regularizers: switching off regularization at the right time can improve performance. In some cases, this can cause an otherwise poorly-performing augmentation to be beneficial.

We hope our findings provide a foundation for continued scientific study of data augmentation. 
\section{Broader Impact}
\paragraph{Data augmentation has the potential to amplify bias}
\*

Data augmentation takes a smaller, potentially biased training set and recycles this as the basis of a larger augmented training program. A central finding of this work is that the success of an augmentation policy varies with the dual metrics of \indistness{} and \diversity{}; as \indistness{} is explicitly model-dependent, it depends on biases present in the model.
This data reuse and model-dependence of successful augmentation suggest the possibility that augmentation may amplify biases in the data or model and warrants future investigation. 

\paragraph{Robust data augmentation can reduce social, environmental, and financial costs}
\*

At its best, data augmentation provides a means for less well-funded or data-rich practitioners to design performant models by supplementing a smaller training data set with additional transformed images.
Commonly-used policies, however, such as those found by AutoAugment~\cite{cubuk2018autoaugment} have relied on expensive brute force searches which cost thousands of GPU-hours, replacing the need for extensive data collections with the need for financially and environmentally expensive compute.
We hope that by understanding the mechanisms behind successful data augmentation we can design guided augmentation policies for new datasets and models, and mitigate the social and financial costs of data collection without undue compute expense. 
\paragraph{Fundamental understanding facilitates impact assessment}
\*

More broadly, the central aim of this work is to better understand the elements driving successful augmentation policies. Truly understanding the conceptual mechanisms at play is crucial in making informed judgements about the impact of data augmentation.

\section*{Acknowledgements}
The authors would like to thank  Alex Alemi, Justin Gilmer, Guy Gur-Ari, Albin Jones, Behnam Neyshabur, Zan Armstrong, and Ben Poole for thoughtful discussions on this work. 


\bibliography{augbib}
\bibliographystyle{unsrtnat}

\newpage
\appendix
\section*{SUPPLEMENTARY MATERIAL}

\section{Training methods}\label{sec:training}
Cifar10 models were trained using code based on AutoAugment code\footnote{available at \texttt{github.com/tensorflow/models/tree/master/research/autoaugment}} using the following choices: 
\begin{enumerate}
    \item Learning rate was decayed following a cosine decay schedule, starting with a value of 0.1
    \item 78050 training steps were used, with data shuffled after every epoch.
    \item As implemented in the AutoAugment code, the WRN-28-2 model was used with stochastic gradient descent and momentum. The optimizer used cross entropy loss with $\ell_2$ weight decay of 0.0005. 
    \item Before selecting the validation set, the full training set was shuffled and balanced such that the subset selected for training was balanced across classes.
    \item Validation set was the last 5000 samples of the shuffled \cifar{} training data. 
    \item Models were trained using Python 2.7 and TensorFlow 1.13 .
\end{enumerate}

A training time of 78k steps was chosen because it showed reasonable convergence with the standard data augmentation of \aug{FlipLR}, \aug{Crop}, and \aug{Cutout} In the clean baseline case, test accuracy actually reached its peak much earlier than 78k steps.

With \cifar{}, experiments were also performed for training dataset sizes of 1024, 4096, and 16384. At smaller dataset sizes, the impact of augmentation and the Switch-off Lift tended to be larger. These results are not shown in this paper.

ImageNet models were ResNet-50 trained using the Cloud TPU codebase\footnote{available at \texttt{https://github.com/tensorflow/tpu/tree/master/models/official/resnet}}. Models were trained for 112.6k steps with a weight decay rate of 1e-4, and a learning rate of 0.2, which was decayed by 10 at epochs 30, 60, and 80. Batch size was set to be 1024. 

\section{Details of augmentation}\label{sec:augdetails}
\subsection{\cifar{}}
\begin{figure}[ht]
 \vskip 0.1in 
 \begin{center}
 \centerline{\includegraphics[width=0.9\columnwidth]{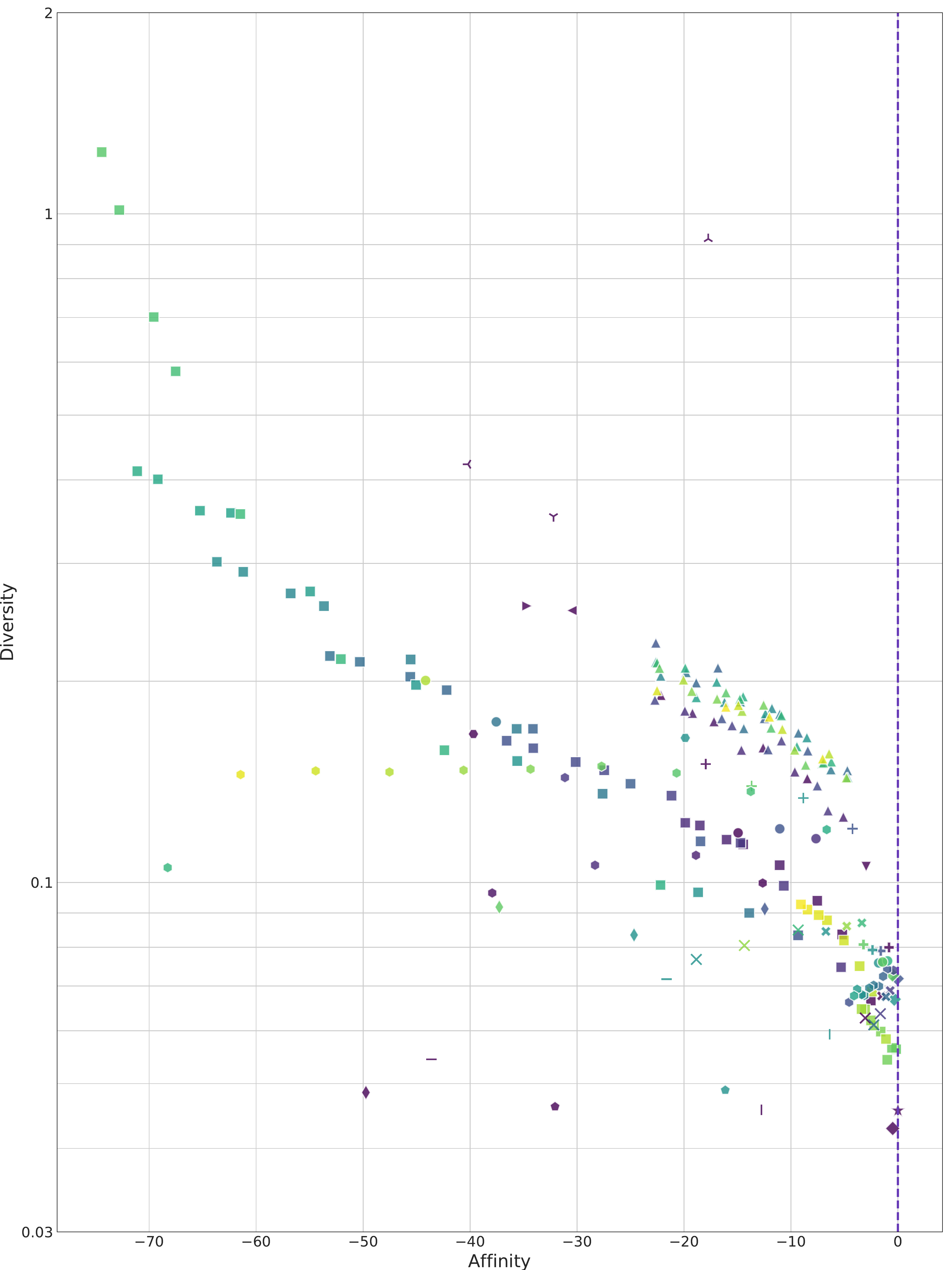}}
 \caption{\cifar{}: Labeled map of tested augmentations on the plane of \indistness{} and \diversity{}. Color distinguishes different hyperparameters for a given transform. Legend is below.}
 \label{fig:appendix_labeledaugs_cf}
 \end{center}
 \vskip -0.1in
\end{figure}

On \cifar{}, both color and affine transforms were tested, as given in the full results (see Sec.~\ref{sec:fullresults}). Most augmentations were as defined in \citet{cubuk2018autoaugment} and additional conventions for augmentations as labeled in Fig.~\ref{fig:appendix_labeledaugs_cf} are defined here. For \aug{Rotate}, {\it fixed} means each augmented image was rotated by exactly the stated amount, with a randomly-chosen direction. {\it Variable} means an augmented image was rotated a random amount up to the given value in a randomly-chosen direction. \aug{Shear} is defined similarly. \aug{Rotate(square)}  means that an image was rotated by an amount chosen randomly from  [0$^\circ$, 90$^\circ$, 180$^\circ$, 270$^\circ$]. 

\aug{Crop} included a padding before the random-location crop so that the final image remained $32\times 32$ in size. The magnitude given for \aug{Crop} is the number of pixels that were added in each dimension. The magnitude given in the label for \aug{Cutout} is the size, in pixels, of each dimension of the square cutout. 

\aug{PatchGaussian} was defined as in \citet{lopes2019improving}, with the patch specified to be contained entirely within the image domain. In Fig.~\ref{fig:appendix_labeledaugs_cf}, it is labeled by two hyperparameters: the size of the square patch (in pixels) that was applied and $\sigma_{max}$, which is the maximum standard deviation of the noise that could be selected for any given patch. Here, ``fixed" means the patch size was always the same. 

Since \aug{FlipLR}, \aug{Crop}, and \aug{Cutout} are part of standard pipelines for \cifar{}, we tested combinations of the three augmentations (varying probabilities of each) as well as these three augmentations plus an single additional augmentation. As in standard processing of \cifar{} images, the first augmentation applied was anything that is not one of \aug{FlipLR}, \aug{Crop}, or \aug{Cutout}. After that, augmentations were applied in the order \aug{Crop}, then \aug{FlipLR}, then \aug{Cutout}.

Finally, we tested the \cifar{} AutoAugment policy~\citep{cubuk2018autoaugment}, RandAugment~\citep{cubuk2019randaugment}, and mixup~\citep{zhang2017mixup}. The hyperparameters for these augmentations followed the guidelines described in the respective papers.   

These augmentations are labeled in Fig.~\ref{fig:appendix_labeledaugs_cf}. 

\begin{figure}[ht]
 \vskip 0.1in 
 \begin{center}
 \centerline{\includegraphics[width=0.9\columnwidth]{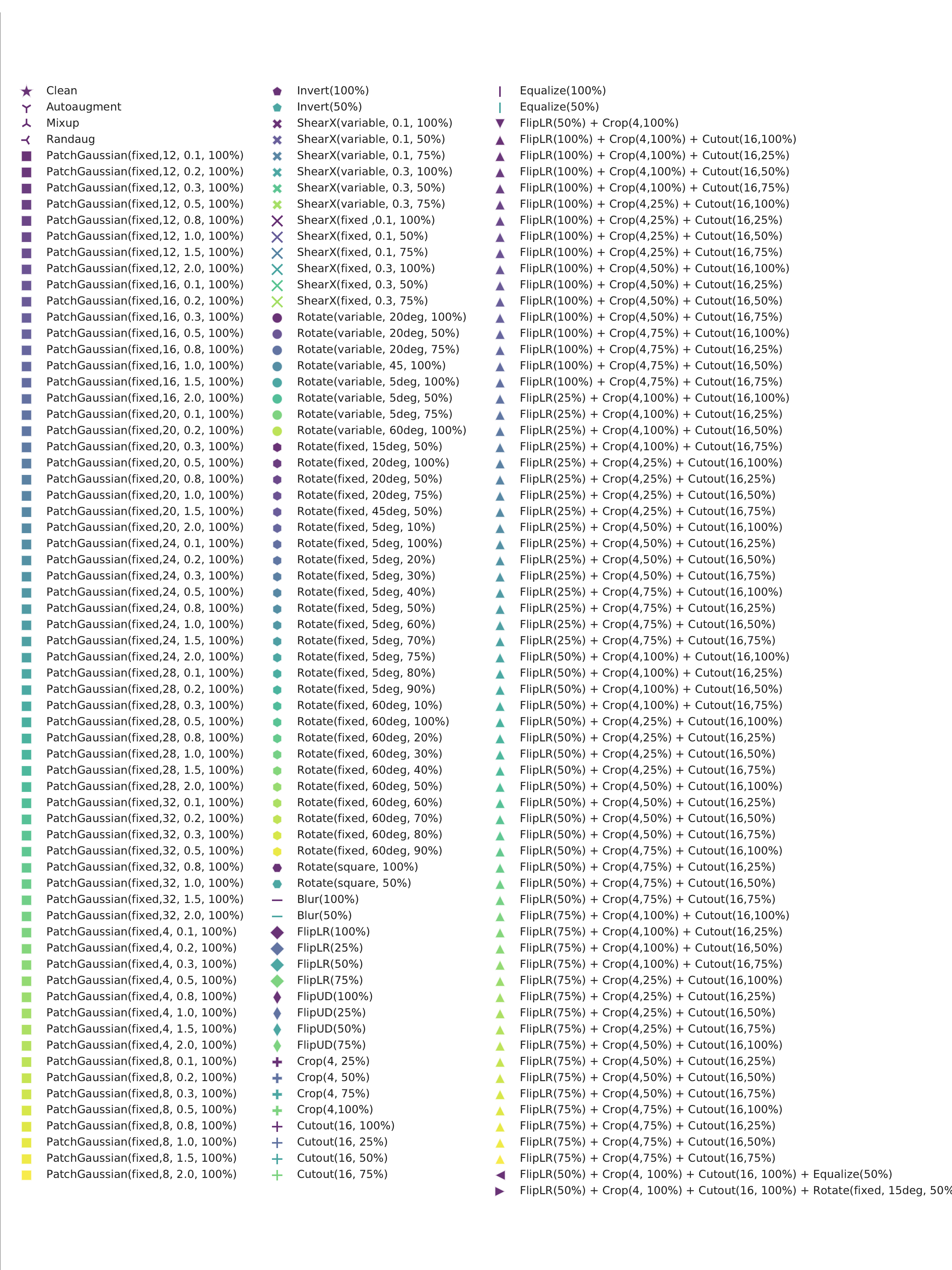}}
    \label{fig:appendix_labeledaugs_legend}
 \end{center}
 \vskip -0.1in
\end{figure}

\subsection{\imagenet{}{}}
On \imagenet{}, we experimented with \aug{PatchGaussian}, \aug{Cutout}, operations from the PIL imaging library\footnote{\url{https://pillow.readthedocs.io/en/5.1.x/}}, and techniques from the AutoAugment code, as described above for \cifar{}. In addition to \aug{PatchGaussian(fixed)}, we also tested \aug{PatchGaussian(variable)}, where the patch size was uniformly sampled up to a maximum size. The implementation here did not constrain the patch to be entirely contained within the image. Additionally, we experimented with \aug{SolarizeAdd}. \aug{SolarizeAdd} is similar to \aug{Solarize} from the PIL library, but has an additional hyperparameter which determines how much value was added to each pixel that is below the threshold. Finally, we also experimented with \aug{Full Gaussian} and \aug{Random Erasing} on \imagenet{}. \aug{Full Gaussian} adds Gaussian noise to the whole image. \aug{Random Erasing} is similar to \aug{Cutout}, but randomly samples the values of the pixels in the patch~\citep{zhong2017random} (whereas \aug{Cutout} sets them to a constant, gray pixel). 

\begin{figure}[ht]
 \vskip 0.1in 
 \begin{center}
 \centerline{\includegraphics[width=0.9\columnwidth]{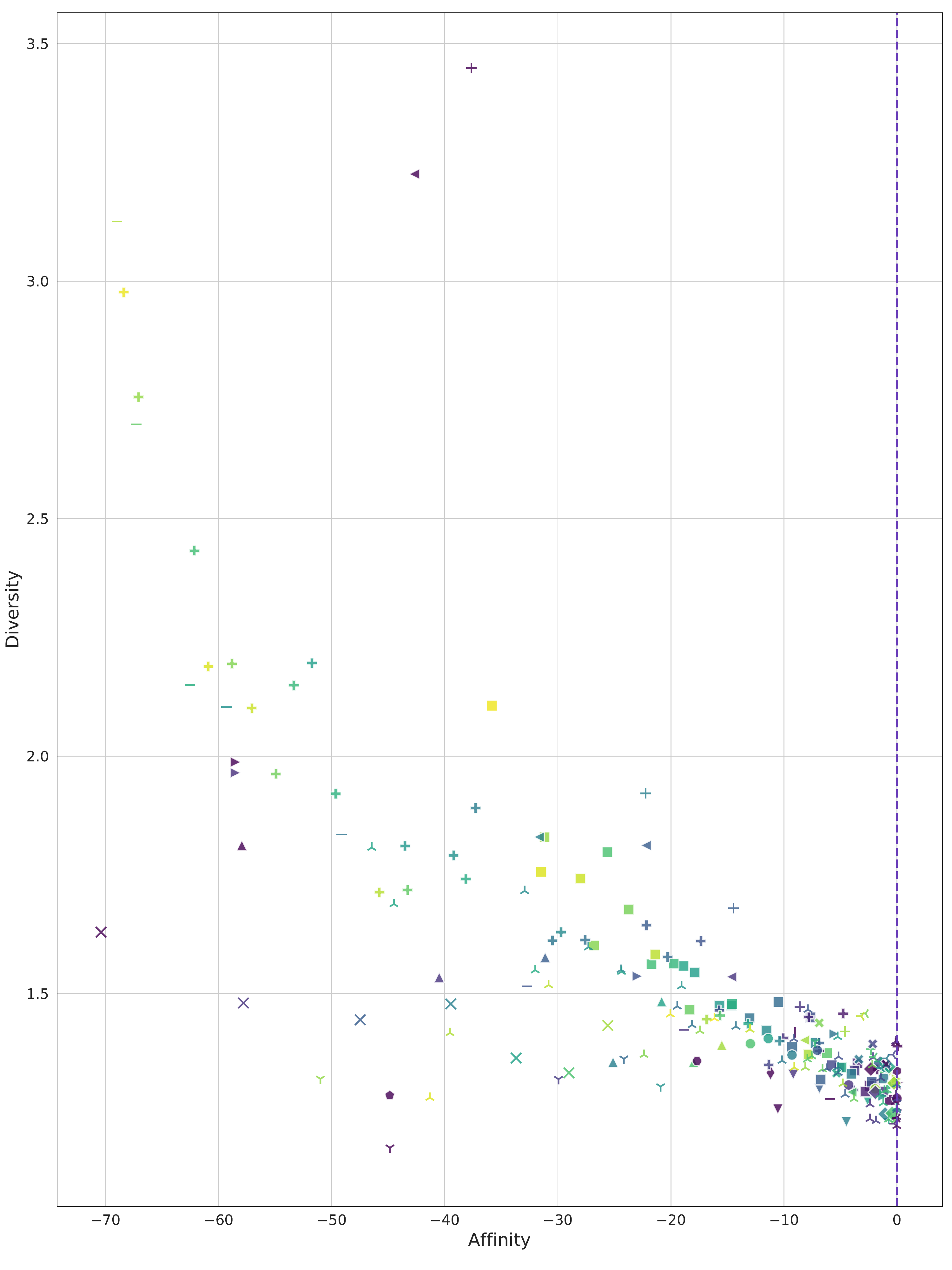}}
 \caption{\imagenet{}: Labeled map of tested augmentations on the plane of \indistness{} and \diversity{}. Color distinguishes different hyperparameters for a given transform. Legend is below.}
 \label{fig:appendix_labeledaugs_in}
 \end{center}
 \vskip -0.1in
\end{figure}

These augmentations are labeled in Fig.~\ref{fig:appendix_labeledaugs_in}.

We note that the gains on \imagenet{} are expected to be small. This is in-line with the magnitude of the gains observed by related works with single transformations~\cite{yin2019afourier}. While combinations of transformations can lead to bigger improvements~\cite{cubuk2019randaugment}, our focus is on understanding single augmentations as a foundation for future work on their combinations.

\begin{figure}[ht]
 \vskip 0.1in 
 \begin{center}
 \centerline{\includegraphics[width=0.9\columnwidth]{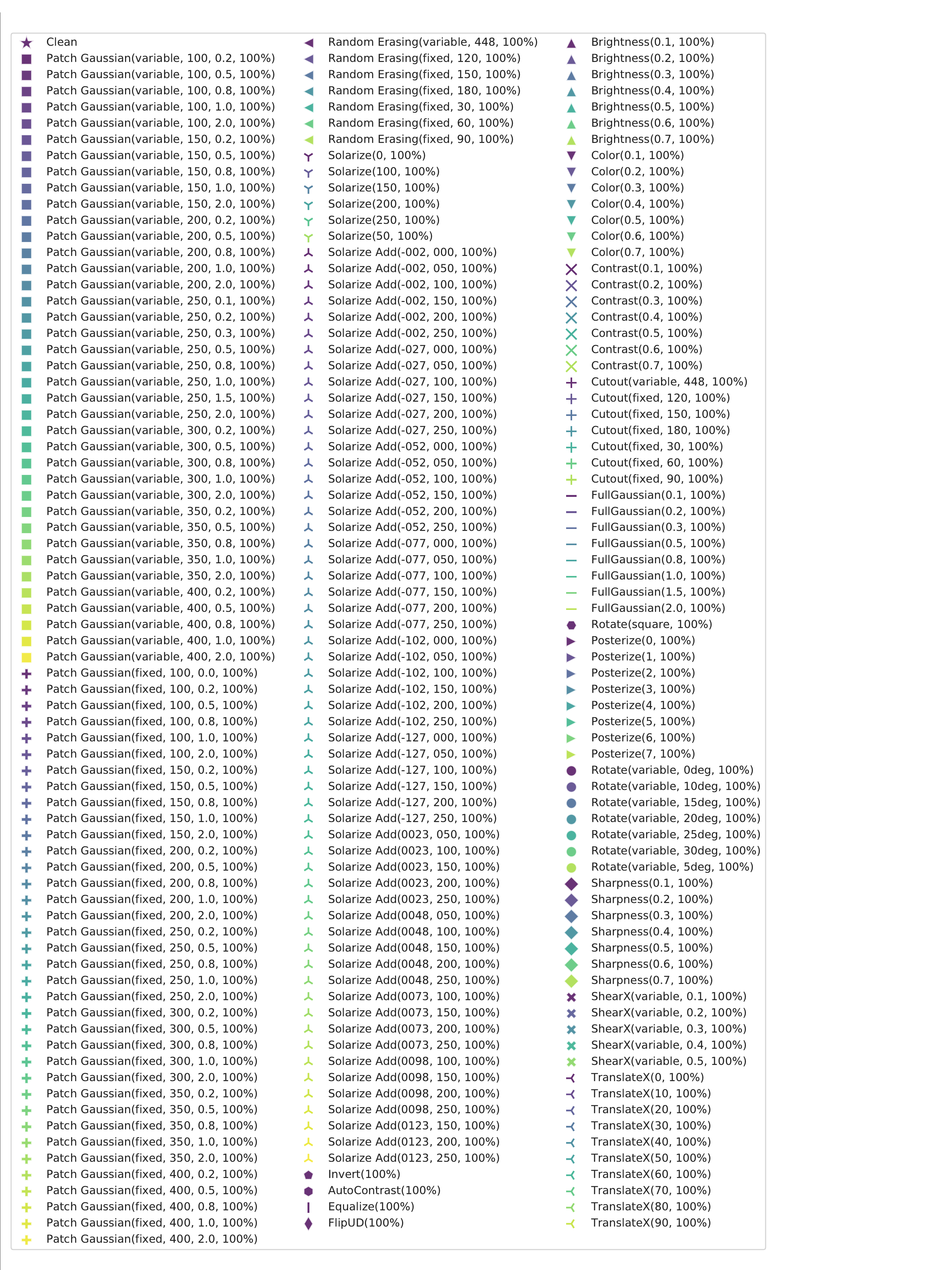}}
 \label{fig:appendix_labeledaugs_legend_in}
 \end{center}
 \vskip -0.1in
\end{figure}

Each augmentation was applied with a certain probability (given as a percentage in the label). Each time an image was pulled for training, the given image was augmented with that probability. 
 
\section{Error analysis}\label{sec:error}
All of the \cifar{} experiments were repeated with 10 different initialization. In most cases, the resulting standard error on the mean (SEM) is too small to show as error bars on plots. The error on each measurement is given in the full results (see Sec.~\ref{sec:fullresults}).
 
\indistness{} and Switch-off Lift both were computed from differences between runs that share the same initialization. For \indistness{}, the same trained model was used for inference on clean validation data and on augmented validation data. Thus, the variance of \indistness{} for the clean baseline is not independent of the variance of \indistness{} for a given augmentation. The difference between the augmentation case and the clean baseline case was taken on a per-experiment basis (for each initialization of the clean baseline model) before the error was computed. 

In the switching experiments, the final training {\it without} augmentation was completed starting from a given checkpoint in the model that was trained {\it with} augmentation. Thus, each switching experiment shared an initialization with an experiment that had no switching. Again, in this case the difference was taken on a per-experiment basis before the error (based on the standard deviation) was computed.

All \imagenet{} experiments shown are with one initialization. Thus, there are not statistics from which to analyze the error.
 
\section{Switching off augmentations}

For \cifar{}, switching times were tested in increments of approximately 5k steps between $\sim 25$k and $\sim 75$k steps. The best point for switching was determined by the final validation accuracy. 

On \imagenet{}, we tested turning augmentation off at 50, 60, 70, and 80 epochs. Total training took 90 epochs. The best point for switching was determined by the final test accuracy. 

The Switch-off Lift was derived from the experiment at the best switch-off point for each augmentation.

For \cifar{}, there are some augmentations where the validation accuracy was best at 25k, which means that further testing is needed to find if the actual optimum switch off point is lower or if the best case is to not train at all with the given augmentation. Some of the best augmentations have a small negative Switch-off Lift, indicating that it is better to train the entire time with the given augmentations. 

For each augmentation, the best time for switch-off is listed in the full results (see Sec.~\ref{sec:fullresults}). 

\section{Diversity metrics}\label{sec:divmetrics}

\begin{figure}[h]
 \vskip 0.1in %
 \begin{center}
 \centerline{\includegraphics[width=0.9\columnwidth]{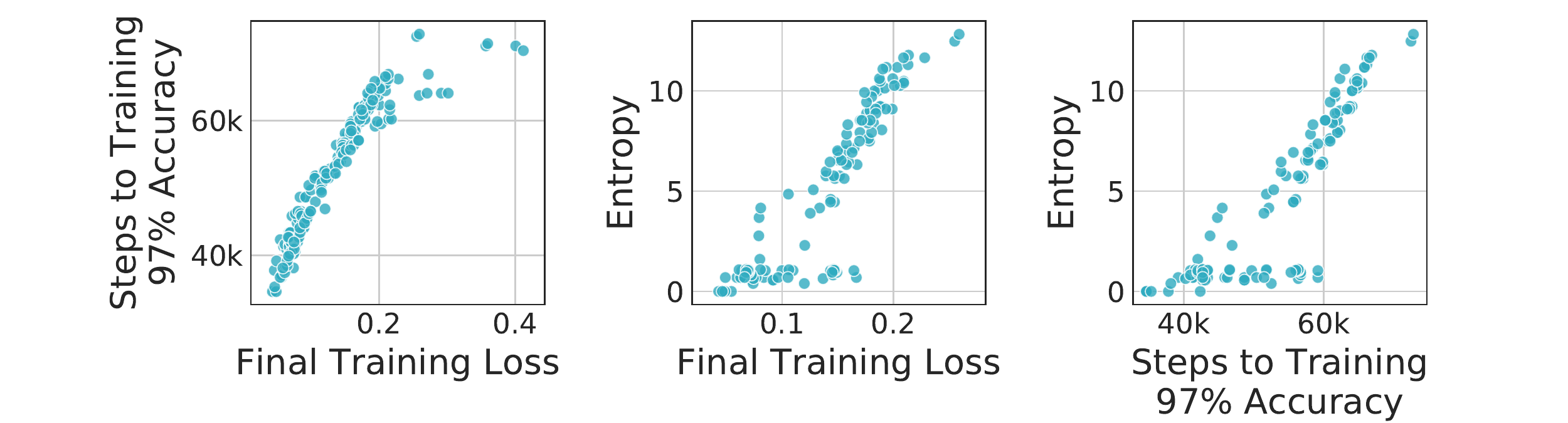}}
 \caption{\cifar{}: Three different diversity metrics are strongly correlated for high entropy augmentations. Here, the entropy is calculated only for discrete augmentations.}
 \label{fig:comparediversitymetrics}
 \end{center}
 \vskip -0.1in
\end{figure}

We computed three possible diversity metrics, shown in Fig.~\ref{fig:comparediversitymetrics}: Entropy, Final Training Loss, Training Steps to Accuracy Threshold. The entropy was calculated only for augmentations that have a discrete stochasticity (such as \aug{Rotate(fixed)} and not for augmentations that have a continuous variation (such as \aug{Rotate(variable)} or \aug{PatchGaussian}). Final Training Loss is the batch statistic at the last step of training. For \cifar{} experiments, this was averaged across the 10 initializations. For \imagenet{}, it was averaged over the last 10 steps of training. Training Steps to Accuracy Threshold is the number of training steps at which the training accuracy first hits a threshold of 97\%. A few of the tested augmentation (extreme versions of \aug{PatchGaussian}) did not reach this threshold in the given time and that column is left blank in the full results. 

Entropy is unique in that it is independent of the model or data set and it is a counting of states. However, it is difficult to compare between discrete and continuously-varying transforms and it is not clear how proper it is to compare even across different types of transforms. 

Final Training Loss and Training Steps to Accuracy Threshold correlate well across the tested transforms. Entropy is highly correlated to these measures for \aug{PatchGaussian} and versions of \aug{FlipLR}, \aug{Crop}, and \aug{Cutout} where only probabilities are varying. For \aug{Rotate} and \aug{Shear} where magnitudes are varying as well, the correlation between Entropy and the other two measures is less clear. 

Building intuition for what \diversity{} means in this case, the Final Training Loss was compared in the case of static augmentation to the case of dynamic augmentation. As shown in Fig.~\ref{fig:fixedaug}, in the case of static augmentation, the \diversity{} was always less than in the typical case of dynamic augmentation. Moreover, across this large range of augmentations, the numerical span of \diversity{} was very small in the case of static augmentation, compared to dynamic augmentation. This suggests that this particular measure of \diversity{} is indeed connected to the number of unique or useful training images that can be created with a given augmentation. In the case of static augmentation, the number of unique images is exactly the same for all augmentations; dynamic augmentations allow for more unique images and both the number and utility of unique images will vary with augmentation.

\section{Comparing \indistness{} to other related measures}\label{sec:supplsewaic}
\begin{figure}[h]
 \vskip 0.1in 
 \begin{center}
 \hfill%
 \begin{subfigure}[\cifar{}]{%
 \includegraphics[width=0.5\columnwidth]{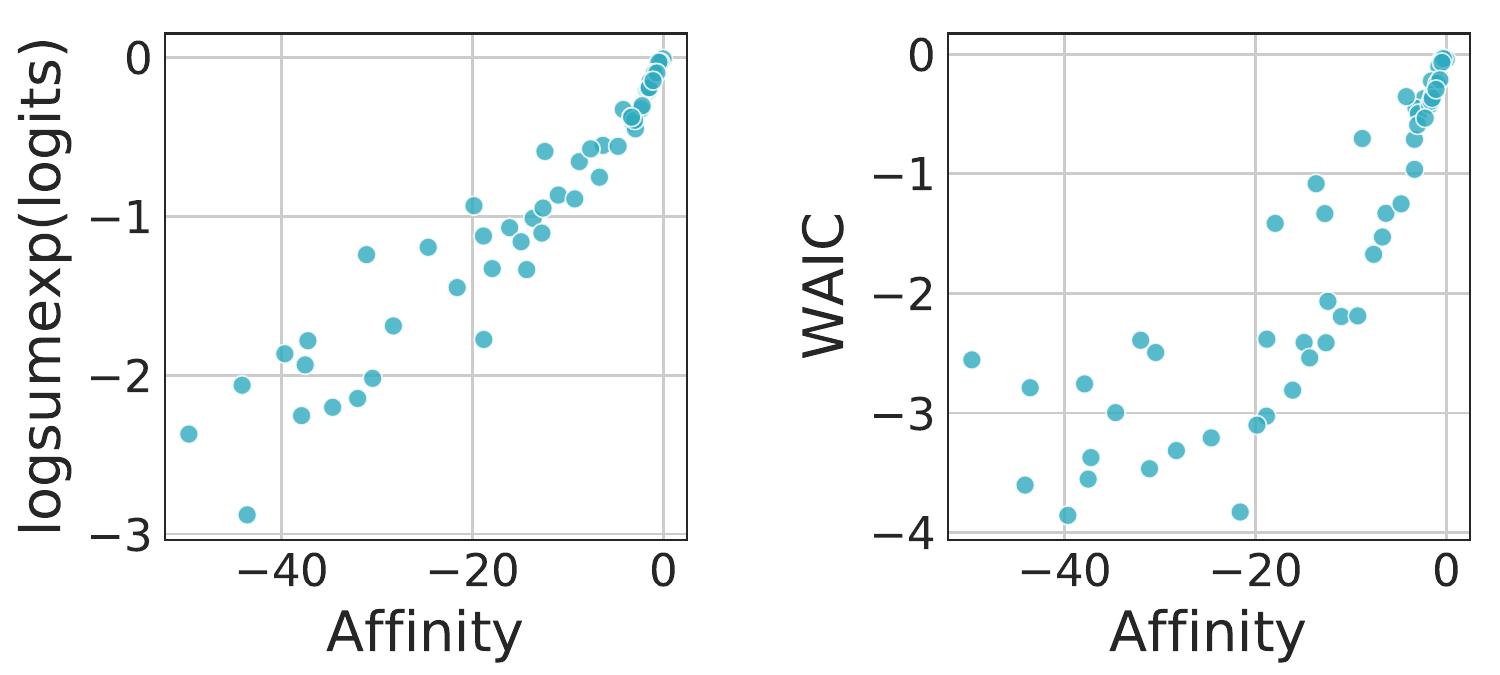}
 }\end{subfigure}%
 \hfill%
 \begin{subfigure}[\imagenet{}]{%
 \includegraphics[width=0.25\columnwidth]{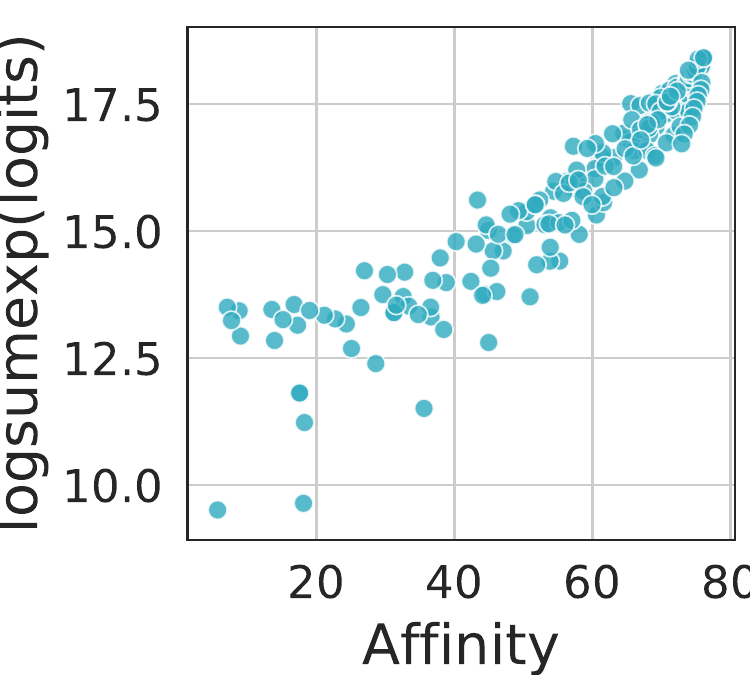}%
 }\end{subfigure}%
 \hfill%
 \caption{\indistness{} correlates with two other measures of how augmented images are related to a trained model's distribution: \aug{logsumexp} of the logits (left, for \cifar{}, and right, for \imagenet{}) is the mean log likelihood for the image. WAIC (middle, for \cifar{}) corrects for a possible bias in that estimate. In all three plots, numbers are referenced to the clean baseline, which is assigned a value of 0.} 
 \label{fig:LSEWAIC}
 \end{center}
 \vskip -0.1in
\end{figure}

We gain confidence in the \indistness{} measure by comparing it to other potential model-dependant measures of distribution shift. In Fig~\ref{fig:LSEWAIC}, we show the correlation between \indistness{} and these two measures: the mean log likelihood of augmented test images\cite{grathwohl2019energyclassifier} (labeled as ``logsumexp(logits)") and the Watanabe–Akaike information criterion (labeled as ``WAIC'') \cite{WAIC}.

Like \indistness{}, these other two measures indicate how well a model trained on clean data comprehends augmented data. 

\section{Full results}\label{sec:fullresults}
The plotted data for \cifar{} and \imagenet{} are given in .csv files uploaded at \url{https://storage.googleapis.com/public_research_data/augmentation/data.zip}. In these .csv files, blank cells generally indicate that a given experiment (such as switching) was not done for the specified augmentation. In the case of the training accuracy threshold as a proxy for diversity, a blank cell indicates that for the given augmentation, the training accuracy did not reach the specified threshold during training.

\end{document}